\documentclass[review,12pt]{elsarticle}

\usepackage{amssymb}
\usepackage{amssymb}
\usepackage{times}  
\usepackage{helvet}  
\usepackage{courier}  
\usepackage{url}  
\usepackage{graphicx}  
\usepackage{dcolumn}
\usepackage{subfigure}
\usepackage{bm}
\usepackage{indentfirst}
\usepackage{amsmath,amsfonts,amssymb}
\usepackage[dvips]{epsfig}
\usepackage{epsfig}
\usepackage{lettrine}
\usepackage[dvips]{graphicx}
\usepackage{multirow}
\usepackage{color}
\usepackage{float}
\usepackage{algorithm}
\usepackage{algpseudocode}
\usepackage{multirow}

\DeclareMathOperator*{\argmin}{argmin}

\journal{Pattern Recognition}

\begin{document}

\begin{frontmatter}



\title{A Unified Weight Learning and Low-Rank Regression Model for Robust Complex Error Modeling}
\author{Miaohua Zhang, Yongsheng Gao, and Jun Zhou}

\address{Institute for Integrated and Intelligent Systems, Griffith University, Australia.}
\begin{abstract}
One of the most important problems in regression-based error model is modeling the complex representation error caused by various corruptions and environment changes in images. For example, in robust face recognition, images are often affected by varying types and levels of corruptions, such as random pixel corruptions, block occlusions, or disguises. However, existing works are not robust enough to solve this problem due to they cannot model the complex corrupted errors very well. In this paper, we address this problem by a unified sparse weight learning and low-rank approximation regression model, which enables the random noises and contiguous occlusions in images
to be treated simultaneously. For the random noise, we define a generalized correntropy (GC) function to match the error distribution. For the structured error caused by occlusions or disguises, we propose a GC function based rank approximation to measure the rank of error matrices. Since the proposed objective function is non-convex, an effective iterative optimization algorithm is developed to achieve the optimal weight learning and low-rank approximation. Extensive experimental results on three public face databases show that the proposed model can fit the error distribution and structure very well, thus obtain better recognition accuracies in comparison with the existing methods.
\end{abstract}

\begin{keyword}
Regression, weight Learning, low-rank approximation, generalized correntropy, robust learning.
\end{keyword}

\end{frontmatter}


\section{Introduction}
\label{}
Regression-based error models can be roughly classified into two categories: the mean square error (MSE) based models and the robust function based ones. The most representative MSE-based approach is the sparse representation classifier (SRC)~\cite{wright2009robust}, which takes advantages of the powerful feature selection ability of sparse representation to learn discriminative features. Deng \textit{et al.}~\cite{deng2017face} extended the SRC by proposing an auxiliary intraclass variant dictionary to characterise the variation between the training and testing images. Huang \textit{et al.} took advantage of the $l_{2,1}$-norm and the label information into consideration to obtain more discriminative features~\cite{huang2013supervised}. However, the performance of these MSE-based methods is significantly deteriorated when data is corrupted by outliers, which is inevitable in real-world applications. Outliers are typically far away from the centre of the data distribution, but MSE-based loss functions assign the same weight to all measures without any discriminative constraints on either severely or slightly corrupted ones when minimizing the representation error, as a consequence, such an equal weight assignment often results in an incorrect sparse solution. Moreover, the MSE-based loss functions assume that the error follows the Gaussian distribution, which is not practical  in real world where usually has more complicated distribution of residuals (errors) caused by non-Gaussian noise or outliers. Therefore, these methods fail to produce sound sparse solutions if the assumption does not hold~\cite{he2011maximum,wang2016correntropy,chen2016generalized}. To overcome these drawbacks, Wright, \textit{et al.}~\cite{wright2009robust} proposed a robust version of SRC to assume that the representation errors or noises are really sparse or almost follow a Laplacian distribution. Then in order to handle illumination variations and strong noise,  Naseem~\textit{et al.}~\cite{naseem2012robust} applied the Huber and Laplacian descriptors. However, these methods assume that the distribution of noise and error distribution is known as prior knowledge, which is difficult to guarantee in practical problems. To overcome this problem, He~\textit{et al.}~\cite{he2011maximum} proposed the CESR algorithm by using a correntropy induced metric to model the representation error. Yang~\textit{et al.}~\cite{yang2012regularized} proposed a regularized robust classifier (RRC) using the local quadratic approximation, and a reweighted least-squares solution is provided. Zheng~\textit{et al.}~\cite{zheng2017iterative} proposed an iteratively reconstrained group sparse classifier (IRGSC) in which an adaptive weight learning procedure is used to give more emphasis on normal image pixels while suppressing the noise and outliers. All the mentioned robust-regression methods had been applied to face recognition and produced promising results.

However, the above algorithms are based on the vector space, and all the image data need to be transferred into vectors before being fed into models. Under this case, the structure of the data (image matrix) is severely destroyed, thus the spatial correlation of the pixels in an image cannot be preserve as much as possible, resulting in inferior recognition performance~\cite{dong2019low,iliadis2017robust,yang2016nuclear}. Recently, researchers pointed out that the error has a specific structure when there are contiguous errors caused by partial occlusion~\cite{iliadis2017robust,xie2017robust,yang2016nuclear}. To make better use of the error structure, \cite{qian2015robust} argued that the error image with occlusion has a low-rank structure, and proposed a low-rank approximation for representing the structure of the error image. The matrix rank minimization, however, is an NP-hard problem and is difficult to be optimized. Yang~\textit{et al.}~\cite{yang2016nuclear} approximated the rank of the error image by the nuclear norm, which significantly improved the performance of face recognition in the presence of occlusions. However, the nuclear norm based rank approximation treats each singular value equally, i.e. it shrinks each singular value with the same threshold regardless of their contribution to the image reconstruction, which leads to a biased estimation. In face recognition, larger singular values of the error image represent the error information corresponding to the occlusion, and smaller ones represent normal image pixels~\cite{iliadis2017robust,xie2017robust,yang2016nuclear}. When performing low-rank approximation on the error image, the error image shall contain as much occlusion information as possible and as less face information as possible. Thus we should give small punishment to large singular values and large punishment to small singular values. Then the resulted error image under a low-rank constraint contains mostly the error information, which means a better approximation. To achieve a better low-rank approximation, some researchers use the non-convex relaxations~\cite{xie2017robust,zheng2019weighted,dong2019low}. Luo ~\textit{et al.}~\cite{luo2016robust} used the Schatten-$p$-norm and obtained a more accurate estimation for error image. Xie~\textit{et al.}~ \cite{xie2017robust} proposed to use a set of non-convex functions to better approximate the low-rank structure of the error image. However, the Schatten-$p$-norm in ~\cite{luo2016robust} treats all the singular values equally, and the convex relaxations in \cite{xie2017robust} may not be an optimal low-rank approximation. Moreover, the existing methods employ different functions for error distribution fitting and structure characterisation, which is cumbersome.

Recently, deep learning (DL) based methods have achieved state-of-the-art performance in various recognition tasks. However, the DL-based methods rely much on huge labeled training data samples for the deep network learning. Once the training data is insufficient, the performance of a DL-based method will degrade severely~\cite{wang2019block}. Moreover, DL-based methods have been verified to have poor performance when the face data is corrupted with complex noises~\cite{dong2019low}.

Considering the weakness of existing methods in fitting the error distribution of complex representation and structure, in this paper, we propose a unified sparse weight learning and low-rank approximation regression model based on the generalized correntropy (GC) to tackle the problem. By choosing different $\alpha$ values in the generalized correntropy, the proposed model can fit various error distribution and approximate the rank of error very well, as shown in Figure 1.
 \begin{figure}[H]
\centering
    \hspace{-0.5cm}
    \vspace{0cm}
   \subfigure[]{\includegraphics[width=0.51\columnwidth]{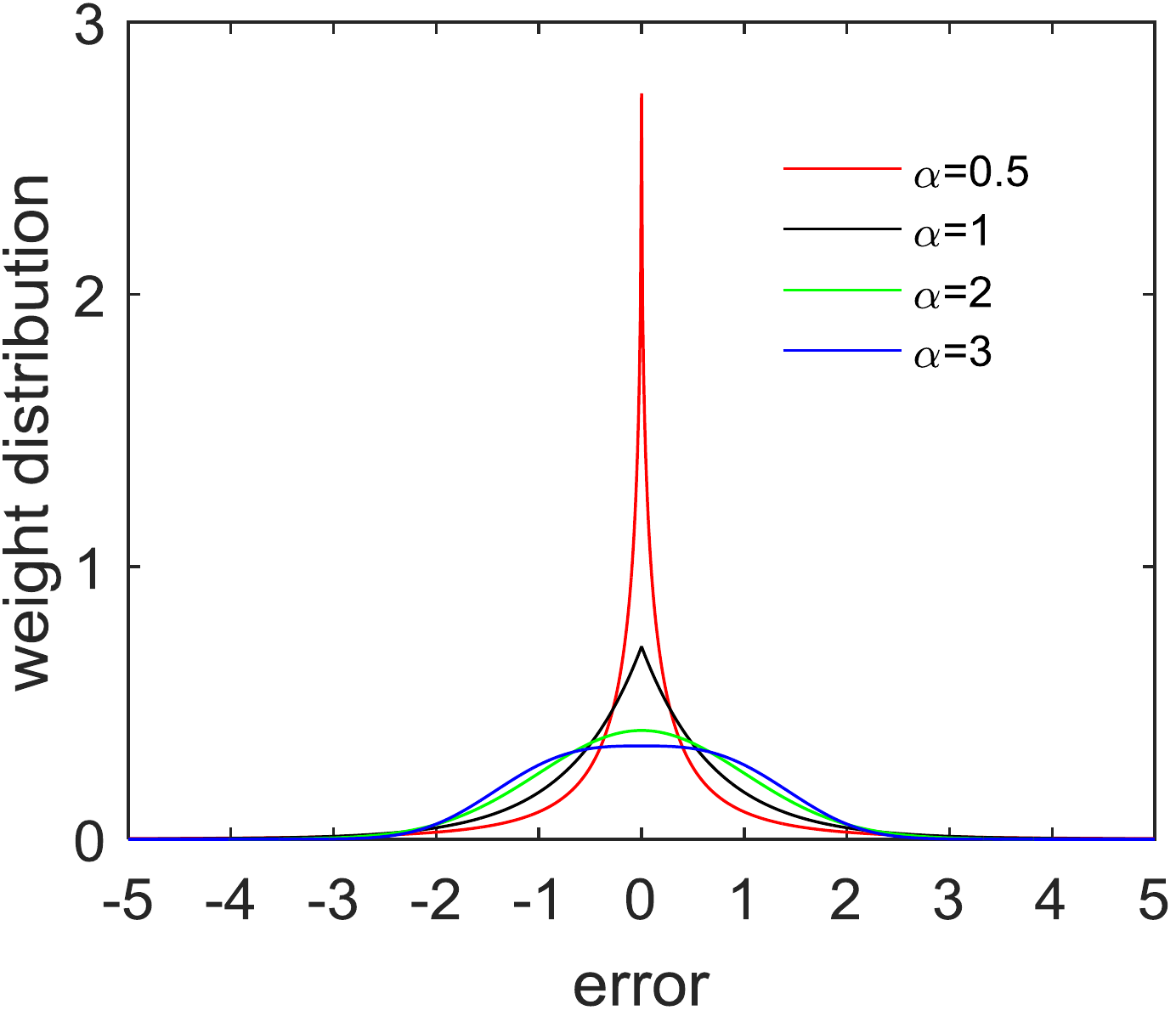}}\;\;
   \subfigure[]{\includegraphics[width=0.49\columnwidth]{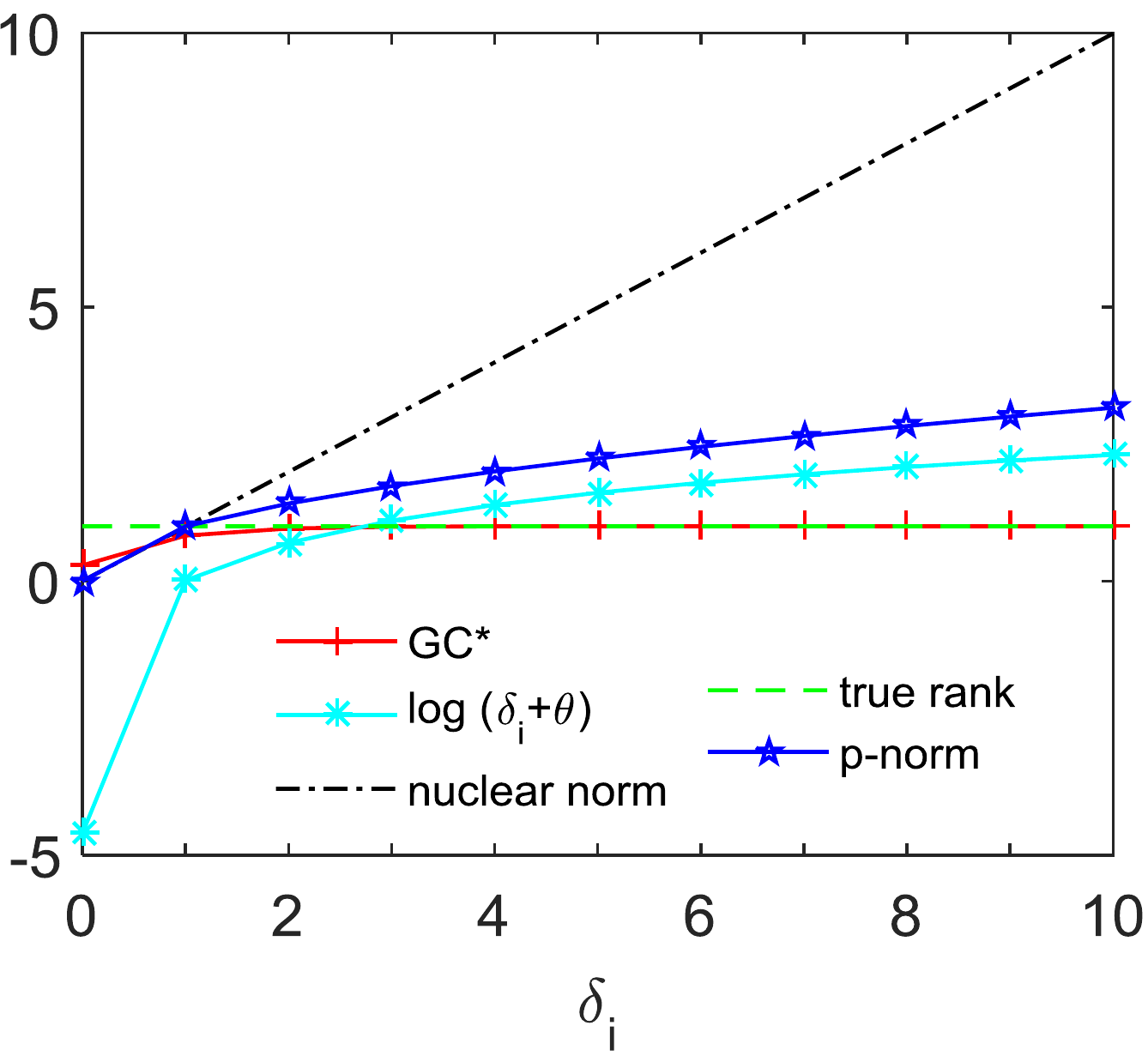}}
    \caption{(a) Weight distribution with different values of parameter $\alpha$ using the GC function; (b) The approximation of different functions for the rank function. Note that the approximated rank of the error by the proposed GC-based rank approximation function (in red) is almost overlapped with the true rank (in green) when $\delta_i$ is greater than 2.}
\end{figure}

The contributions of this paper are summarized as follows.
\begin{itemize}
\item The proposed algorithm is the first attempt to use only one function to address both error distribution fitting and structure estimation, which provides a new theory and framework for regression-based complex error modeling.
 \item To overcome the vulnerability of the existing methods in fitting error distribution of complex representation, we propose a generalized-correntropy-based weight learning theory which can better fit the variation of illumination, expressions, poses, noise, and occlusions in face images.
\item A new and more accurate low-rank approximation estimator based on the generalized correntropy is proposed for contiguous error structure estimation.
\item Since the proposed weight learning and low-rank approximation function are non-convex, we propose an optimization scheme based on the majorization minimization (MM) and alternating direction method of multipliers (ADMM) to solve it with guaranteed convergence.
 \end{itemize}
The remainder of this paper is organized as follows: Related works are introduced in Section 2. The proposed method, including the definition of new weight learning and low-rank approximation function, is introduced in Section 3. The specifically designed optimization method for the proposed objective function is included in Section 4. In Section 5, we analyse the computational complexity and the convergence property of the proposed algorithm. Section 6 gives experimental results of our algorithm compared with the benchmark methods. Finally, conclusions are drawn in Section 7.

\section{Related Works}
 \subsection{Robust Weight Learning}
In this section, we briefly the review robust weighting learning and low-rank approximation, which forms the foundation of our approach.

 He ~\textit{et al.}~\cite{he2011maximum} proposed a correntropy induced metric (CIM) based loss function for  robust  face  recognition. They adaptively learn a weight for the representation error, by which the larger errors corresponding to the noise and outliers receive smaller weights (larger penalty), while the smaller errors receive larger weights (smaller penalty). Given a query image vector $\mathbf{y}\in \mathbb{R}^M$, and a training dataset $\mathbf{D}\in R^{M\times N}$, the CIM-based loss function and weight estimators are defined as follows.
 \begin{equation}
\begin{aligned}
    &\hat{J}=\underset{x,w}{\max}~\sum_{j=1}^M (w_j(y_j-\sum_{i=1}^N d_{i,j}x_i)^2 -\phi(w_j))-\lambda \sum_{i=1}^N x_i,\\
    &~~~~~~~~~~~~~~~ \text{subject to (s.t.)} ~~~~x_i\geq 0,~~i=1,\cdots, N,~~ j=1, \cdots, M.
    \end{aligned}
\end{equation}

\noindent where the weight is calculated by
\begin{equation}
    \begin{aligned}
     w_j=-g(y_j -\sum_{i=1}^N d_{ij}x_i ), ~~i=1,\cdots, N,~~ j=1, \cdots, M.
    \end{aligned}
\end{equation}
where $g(\cdot)$ is the Gaussian function.

Considering that the Gaussian function in \cite{he2011maximum} is not robust enough to match the error when there are heavy noises and large occlusions, Iliadis~\textit{et al.}~\cite{iliadis2017robust} and Yang~\textit{et al.}~\cite{yang2012regularized} proposed to use the logistic function as a weight descriptor to match the error distribution as
\begin{equation}
    \begin{aligned}
     w_j=\frac{\text{exp}(-\gamma e_j^2+\gamma\theta)}{1+\text{exp}(-\beta e_j^2+\beta\theta)},~~~~ j=1, \cdots, M.
    \end{aligned}
\end{equation}
where $e_j$ denotes the representation error.

Different from learning weight using a specific function as in \cite{he2011maximum,iliadis2017robust}, \cite{zheng2017iterative} proposed an iterative procedure to adaptively learn the weight by solving a constrained sparse learning problem. Their model is defined as follows:
\begin{equation}
    \begin{aligned}
     \underset{\mathbf{w}^T \mathbf{1}=1,\mathbf{w}}{\argmin}~\frac{1}{2}\|\sqrt{\mathbf{w}}\odot(\mathbf{y}-\mathbf{D}\mathbf{x})\|_2^2+\gamma\|\mathbf{w}\|_2^2,
    \end{aligned}
\end{equation}
where $\odot$ denotes the Hadamard product, and $\mathbf{w}$ is $[\omega_1, \cdots, \omega_j, \cdots, \omega_M]$ with each $\omega_j$ updated by 
\begin{equation}
    \begin{aligned}
     w_j=(-\frac{\mathbf{d}}{2\gamma}+\eta)_+, ~~~~ j=1, \cdots, M,
    \end{aligned}
\end{equation}
where $\mathbf{d}$ is a vector consisting of representation error, i.e., $\mathbf{d}=[e_1^2, \cdots, e_j^2, \cdots,e_M^2]$, $\eta$ is the Lagrangian multiplier, and $\gamma$ is a tunable parameter for the $l_2$-norm regularization. Here $(\cdot)_+$ is a threshold function that sets the negative values to zeros while keeping the positive ones.

\subsection{Low-Rank Approximation}
 Both \cite{iliadis2017robust} and \cite{yang2016nuclear} used the nuclear norm to approximate the low-rank structure of the error image in the presence of contiguous occlusions. Let matrix $\mathbf{E}$ be the error image, the rank-constrained error can be calculated as
 \begin{equation}
    \begin{aligned}
 \underset{\hat{\mathbf{E}}}{\min}~\frac{1}{2}\|\hat{\mathbf{E}}-\mathbf{E}\|_F^2+\lambda\|\hat{\mathbf{E}}\|_*,
  \end{aligned}
\end{equation}
where $\|\hat{\mathbf{E}}\|_*$ is the nuclear norm of $\hat{\mathbf{E}}$, which is defined as the sum of its singular values. Then the optimal low-rank constrained error image is given by
\begin{equation}
    \begin{aligned}
    \hat{\mathbf{E}}^*=\mathbf{U}\mathbf{S}\mathbf{V}^T,
    \end{aligned}
\end{equation}
where $\mathbf{U}$ and $\mathbf{V}$ are respectively the left and right singular matrices of $\mathbf{E}$, $\mathbf{S}=\text{sign}(\delta_{i})\text{max}(0,|\delta_{i}|-\lambda)$, and $\delta_{i}$ ($i=1,2,\ldots,N$) are the singular values of $\mathbf{E}$.

The nuclear norm based low-rank approximation treats each singular value equally regardless their contributions to the error image. Then Xie~\textit{et al.}~\cite{xie2017robust} proposed to use the non-convex function to better approximate the low-rank structure of the error image. Their robust low-rank model is defined by
\begin{equation}
    \begin{aligned}
    \underset{\hat{\mathbf{E}}}{\min}~\frac{1}{2}\|\hat{\mathbf{E}}-\mathbf{E}\|_F^2+\lambda\|\hat{\mathbf{E}}\|_{\omega,*},
    \end{aligned}
\end{equation}
where the second term is used for rank approximation and can be relaxed by some non-convex functions including $l_p$-norm, $\text{log-sum}$, $\text{atan}$, and $\text{log-exp}$ functions. Then the optimal error matrix with low-rank property can be calculated by
\begin{equation}
    \begin{aligned}
    \hat{\mathbf{E}}^*=\mathbf{U}\mathbf{S}_{\omega,*}\mathbf{V}^T,
\end{aligned}
\end{equation}
where $\mathbf{U}$ and $\mathbf{V}$ are the left and right singular matrices of $\mathbf{E}$, and $\mathbf{S}_{\omega,*}$ is the weighted Singular Value Thresholding (SVT) operator,
\begin{equation}
    \begin{aligned}
    & \mathbf{S}_{\omega,*}=\text{diag}(\text{max}(\delta_i-\omega_i\lambda,0)),~~i=1,2,\cdots, M_2,
\end{aligned}
\end{equation}
where $\delta_i$ is the singular value of $\mathbf{E}$, and $\omega_i$ is a weight controlling the shrinkage level of each singular value.

\section{Proposed Method}
As described previously, the existing methods cannot effectively fit the complex representation error, and most of these methods adopt two separate functions for pixel and structural corruptions estimation that is cumbersome. We proposed a unified weight learning and low-rank approximation (UWLLA) regression model to solve these problems. First, a non-convex generalized correntropy (GC) function is proposed to fit the complex error distribution and approximate the low-rank structure of the error matrix. Then an optimization algorithm based on the majorization minimization (MM) theory is developed to solve the non-convex objective function.

 \subsection{Generalized Correntropy}
 Motivated by the successfully application of the non-convex function in image processing ~\cite{xie2017robust,xie2019hyperspectral} and GC function.
  We propose a nonconvex function $f_{\text{GC}}(x)$ based on the GC function to measure the contribution of each point $\mathbf{e}\in R^m$ as
 \begin{equation}
\begin{aligned}
&f_{\text{GC}}(A-B)=\frac{1}{2}E\left[||\varphi_{\alpha,\beta}(A)-\varphi_{\alpha,\beta}(B)||_{\mathcal{H}}^2\right]\\
&=\frac{1}{2}E[\langle\varphi_{\alpha,\beta}(A),\varphi_{\alpha,\beta}(A)\rangle+\langle\varphi_{\alpha,\beta}(B),\varphi_{\alpha,\beta}(B)\rangle \\
&~~~~-2\langle\varphi_{\alpha,\beta}(A),\varphi_{\alpha,\beta}(B)\rangle]\\
&=E[(G_{\alpha,\beta}(0)-G_{\alpha,\beta}(\mathbf{e}))],
\end{aligned}
\end{equation}
where $A$ and $B$ are two random variables, $\mathbf{e}=A-B$, $E[x]$ is the expectation of $x$, $\varphi_{\alpha,\beta}(\cdot)$ denotes a nonlinear mapping which transforms its argument into a high-dimensional Hilbert space~\cite{chen2016generalized}, $\|\cdot\|_{\mathcal{H}}$ denotes transformation operation in Hilbert space, and $G_{\alpha,\beta}$ is the Generalized Gaussian Density (GGD) function given by
 \begin{equation}
  \begin{aligned}
  &~ G_{\alpha,\beta}(e)=\frac{\alpha}{2\beta\Gamma(1/\alpha)}\text{exp}\left(-\left|\frac{e}{\beta}\right|^\alpha\right)\\
  &~~~~~~~~~~~~~~=\gamma_{\alpha,\beta}\text{exp}\left(-\lambda\left|e\right|^\alpha\right).
  \end{aligned}
  \end{equation}

 Here, $\alpha>0$ and $\beta>0$ are the parameters of GGD indicating the peak and width of the probability density function. $\Gamma(z)=\int_0^\infty e^{-t}t^{z-1}dt, (z>0)$ is the gamma function. $\lambda=1/\beta^\alpha$ and $\gamma_{\alpha,\beta}=\alpha/(2\beta\Gamma(1/\alpha))$ are the kernel parameter and the normalization constant, respectively. Obviously, the Gaussian function is just a special case of the generalized Gaussian density function when $\alpha$ is $2$. When $\alpha$ is 1, equation (12) becomes the Laplacian distribution. We plot the GGD distributions with several shape parameters in Figure 1(a) which shows that smaller values of $\alpha$ give heavier tails (sharper distributions). When $\alpha\rightarrow \infty$, the GGD is close to the uniform distribution, while when $\alpha\rightarrow 0_+$, the GGD approaches an impulse function. Thus, owing to the flexibility of shape parameter selection, the GGD function can match the errors of different distributions very well.

Based on the above analysis, we give two definitions for the weight learning and low-rank approximation here.

\noindent\textbf{Definition 1:} Given the representation error vector $\mathbf{e}\in R^m$, the weight learning function can be defined as
\begin{equation}
f_{\text{GC}}(\mathbf{e})=\gamma_{\alpha,\beta}(1-\text{exp}(-\lambda|\frac{\mathbf{e}}{\|\mathbf{e}\|_{\infty}}|^{\alpha})),
 \end{equation}
where $\|\mathbf{e}\|_{\infty}$ is used to normalize the representation error and ensure the errors to be in the same scale. We can see that the function $f_{\text{GC}}(\cdot)$ treats each entry adaptively. By choosing different $\alpha$, Definition 1 can fit many complex distributions, the error fitting curve in Figure 1(a) shows that the GC function can fit different levels of error, especially the smaller residual errors. In Definition 1, the large representation error will be given a large penalty, making the representation procedure less affected by large corruptions.

\noindent\textbf{Definition 2:} Given an error matrix $\mathbf{E}$, the rank approximation based on the GC-function is defined as follows:
\begin{equation}
f_{\text{GC}}(\sigma(\mathbf{E}))=\gamma_{\alpha,\beta}(1-\text{exp}(-\lambda|\sigma(\mathbf{E})|^{\alpha})),
 \end{equation}
 where $\mathbf{E}\in R^{m_1\times m_2}$ is the matrix form of the representation error vector, and $\sigma(\mathbf{E})$ represents the singular values of $\mathbf{E}$. Definition 2 shrinks the larger singular value less and the smaller value more, which provides discriminative measurements for the pixels in the error image, and thus the low-rank approximation will be more accurate. The rank approximations using different functions are plotted in  Figure 1(b) which shows that the GC-function based rank approximation in Definition 2 has a better approximation of the true rank than other functions.
\subsection{The Proposed UWLLA}
Motivated by the advantages of the GC-function in learning discriminative weights and in approximating rank for errors in Definitions 1 and 2, we consider using $f_{\text{GC}}(\mathbf{e})$ as the weight learning function, and $f_{\text{GC}}(\sigma(\mathbf{E}))$ as surrogate function for matrix rank approximation to learn more robust features in the presence of noises, outliers, and occlusions. The proposed UWLLA model is defined as
 \begin{equation}
 \begin{aligned}
 &\min_{\mathbf{w},\mathbf{x}} f_{\text{GC}}(\mathbf{e})+\lambda_1 f_{\text{GC}}(\sigma(\mathbf{E})) +\lambda_2 v(\mathbf{x}),\\
 & ~~~~~~~\text{s.t.}~~\mathbf{y}-\mathbf{D}\mathbf{x}=\mathbf{e}, ~~~\mathbf{E}=TM(\mathbf{e}).
 \end{aligned}
 \end{equation}
 where $\lambda_1>0$ and $\lambda_2>0$ are regularization parameters used to control the tradeoff between the constraints of sparsity and matrix rank, and $TM(\mathbf{e})$ means transforming the vector $\mathbf{e}$ to the matrix form $\mathbf{E}$. However, the weight learning and low-rank approximation functions in objective function (15) are non-convex and difficult to optimize. Thus we design an optimization algorithm for equation (15) in the following.

\section{Optimization by Majorization Minimization}
In recent years, the majorization minimization (MM) theory has been verified to have superior ability in solving non-convex and non-smooth problems in research field of computer vision, machine learning, and signal processing~\cite{liang2020correlation}\cite{schultz2018nonsmooth} \cite{chopra2010total}\cite{yang2016multi}\cite{ren2019sinusoidal}\cite{marnissi2020majorize}.
In this section, we present an MM optimization strategy to solve the proposed objective function in (15). Instead of solving the complicated nonconvex optimization problem directly, the MM technique solves a set of convex surrogate optimization problems~\cite{fan2020min}\cite{sun2016majorization}. Specifically, for an given optimization problem $\min_{\mathbf{x}\in\mathcal{X}} f(\mathbf{x})$ where $\mathcal{X}$ is the feasible set, $f(\mathbf{x})$ is the non-convex objective function which is difficult to solve directly.
The MM replaces the original function with its upper-bound surrogate function in the majorization step and then minimize the resulted function in the minimization step, which can be described in the following two steps.

\noindent 1) \textit{Majorization}: we first define the majorization surrogate function (upper bound) $f(\mathbf{x}|\mathbf{x}_t)$ for the non-convex objective function $f(\mathbf{x})$ at $\mathbf{x}_t$,
\begin{equation}
f(\mathbf{x}|\mathbf{x}_t)\geq f(\mathbf{x})+c_t, ~~~\forall \mathbf{x}\in \mathcal{X}.
\end{equation}
The difference between $f(\cdot|\mathbf{x}_t)$ and $f(\mathbf{x})$ is minimized at $\mathbf{x}_t$, and $c_t=f(\mathbf{x}_t|\mathbf{x}_t)-f(\mathbf{x}_t)$.

\noindent 2) \textit{Minimization}: in the minimization step, the optimization solution to the surrogate function is solved by
\begin{equation}
\mathbf{x}_{t+1}=\underset{\mathbf{x}\in \mathcal{X}}{\argmin} ~f(\mathbf{x}|\mathbf{x}_t).
\end{equation}
As in~\cite{sun2016majorization} , the first-order Taylor expansion of $f(\mathbf{x})$ is used as a surrogate function as follows:
\begin{equation}
f(\mathbf{x})\leq f(\mathbf{x}_t)+f'(\mathbf{x}_t)(\mathbf{x}-\mathbf{x}_t)=f(\mathbf{x}|\mathbf{x}_t).
\end{equation}
Then $f(\mathbf{x})$ can be upper-bounded as
\begin{equation}
f(\mathbf{x})\leq f(\mathbf{x}_t)+f'(\mathbf{x}_t)\mathbf{x}+c_t,
\end{equation}
where $c_t$ is a constant. We then apply the MM procedures for the weight learning and low-rank approximation non-convex functions in (15) step by step.
\subsection{Majorization Procedure}
Let $f_{\text{GC}}(\mathbf{e}|\mathbf{e}_t)$ and $f_{\text{GC}}(\sigma(\mathbf{E})|\sigma(\mathbf{E}_t))$ be the upper bound surrogate function for weight learning function $f_{\text{GC}}(\mathbf{e})$ and low-rank approximation function $f_{\text{GC}}(\mathbf{E})$, according to MM theory described above, the majorization functions for the weight learning can be defined by
\begin{equation}
f_{\text{GC}}(\mathbf{e})\leq f_{\text{GC}}(\mathbf{e}_t)+f'_{\text{GC}}(\mathbf{e}_t)\left(\mathbf{e}-\mathbf{e}_t\right),
\end{equation}
and for the low-rank approximation is
\begin{equation}
f_{\text{GC}}(\sigma(\mathbf{E}))\leq f_{\text{GC}}(\sigma(\mathbf{E}_t))+f'_{\text{GC}}(\sigma(\mathbf{E}_t))\left(\sigma(\mathbf{E})-\sigma(\mathbf{E}_t)\right).
\end{equation}
where $\mathbf{e}_t$ and $\mathbf{E}_t$ denotes the error vector and matrix calculated in the previous iteration. $f_{\text{GC}}'(\mathbf{e}_t)$ and $f_{\text{GC}}'(\sigma(\mathbf{E}_t))$ represent the first order derivative of $f_{\text{GC}}(\mathbf{e}_t)$ and $f_{\text{GC}}(\sigma(\mathbf{E}_t))$. Then $f_{\text{\text{GC}}}(\mathbf{e})$ and $f_{\text{\text{GC}}}(\sigma(\mathbf{E}))$ can be upper-bounded as
\begin{equation}
\begin{aligned}
&f_{\text{GC}}(\mathbf{e})\leq f_{\text{GC}}(\mathbf{e}_t)+f'_{\text{GC}}(\mathbf{e}_t)\mathbf{e}+c=f_{\text{GC}}(\mathbf{e}|\mathbf{e}_t),\\
&f_{\text{GC}}(\sigma(\mathbf{E}))\leq f_{\text{GC}}(\sigma(\mathbf{E}_t))+f'_{\text{GC}}(\sigma(\mathbf{E}_t))\sigma(\mathbf{E})+c\\
&~~~~~~~~~~~~~~~~~~=f_{\text{GC}}(\sigma(\mathbf{E}|\sigma(\mathbf{E}_t))).
\end{aligned}
\end{equation}

\subsection{Minimization Procedure}
Based on the above analysis, minimizing the objective function in equation (15) can be solved by minimizing the following surrogate function
\begin{equation}
\begin{aligned}
& \underset{\mathbf{e},\mathbf{x},\mathbf{h}}{\argmin}~~~f_{\text{GC}}(\mathbf{e}|\mathbf{e}_t)+\lambda_1f_{\text{GC}}(\sigma(\mathbf{E})|\sigma(\mathbf{E_t}))+\lambda_2v(\mathbf{h}), \\
& \text{s.t.}~~~ \mathbf{y}-\mathbf{D}\mathbf{x}=\mathbf{e},~~~\mathbf{E}=TM(\mathbf{e}), ~~~\mathbf{x}=\mathbf{h}.
\end{aligned}
\end{equation}
 Here the constraint $\mathbf{x}=\mathbf{h}$ is introduced to ensure that the optimized coefficient $\mathbf{x}$ is nonnegative. Problem (23) can be efficiently solved by ADMM technique which  breaks the objective function into smaller pieces and obtains an approximation solution with fast convergence~\cite{boyd2011distributed,hu2019doubly,piao2019double}. The augmented Lagrangian function of equation (23) is given by
 \begin{equation}
 \begin{aligned}
 &\mathcal{L}(\mathbf{e}, \mathbf{x},\mathbf{h},\mathbf{v}_1,\mathbf{v}_2)=f_{\text{GC}}(\mathbf{e}|\mathbf{e}_t)+\lambda_1f_{\text{GC}}(\sigma(\mathbf{E})|\sigma(\mathbf{E_t}))\\
 &+\lambda_2v(\mathbf{h})+\mathbf{v}_1^T(\mathbf{y}-\mathbf{Dx-e})+\frac{\rho_1}{2}\|\mathbf{y}-\mathbf{Dx-e}\|_2^2\\
 &~+\mathbf{v}_2^T(\mathbf{x-h})+\frac{\rho_2}{2}\|\mathbf{x-h}\|_2^2,
 \end{aligned}
 \end{equation}
 where $\rho_1$ and $\rho_2$ are positive penalty parameters, and $\mathbf{v}_1$ and $\mathbf{v}_2$ are the dual variables. The optimal parameters can be updated by the following ADMM procedure.
\begin{equation}
\begin{aligned}
&\mathbf{e}_{t+1}=\underset{\mathbf{e}}{\argmin}~\mathcal{L}(\mathbf{e}, \mathbf{x}_t,\mathbf{h}_t,\mathbf{v}_{1,t},\mathbf{v}_{2,t}),\\
&\mathbf{h}_{t+1}=\underset{\mathbf{z}}{\argmin}~\mathcal{L}(\mathbf{e}_{t+1},\mathbf{x}_t,\mathbf{h},\mathbf{v}_{1,t},\mathbf{v}_{2,t}),\\
&\mathbf{x}_{t+1}=\underset{\mathbf{x}}{\argmin}~\mathcal{L}(\mathbf{e}_{t+1},\mathbf{x},\mathbf{h}_{t+1},\mathbf{v}_{1,t},\mathbf{v}_{2,t}),\\
&\mathbf{v}_{1,t+1}=\mathbf{v}_{1,t}+\rho_1(\mathbf{y}-\mathbf{D}\mathbf{x}_{t+1}-\mathbf{e}_{t+1}),\\
&\mathbf{v}_{2,t+1}=\mathbf{v}_{2,t}+\rho_2(\mathbf{x}_{t+1}-\mathbf{h}_{t+1}),\\
\end{aligned}
\end{equation}
where $t$ is the optimization iteration. In the following, we alternatively solve all the variables in equation (25) by solving one variable at a time while fixing others.
\subsubsection{Updating $\mathbf{e}_{t+1}$}
The optimal $\mathbf{e}_{t+1}$ can be updated by solving the following problem:
\begin{equation}
\begin{aligned}
\!\!\!\!\mathbf{e}_{t+1}=&\:\underset{\mathbf{e}}{\argmin}~~f_{\text{GC}}(\mathbf{e}|\mathbf{e}_t)+\lambda_1f_{\text{GC}}(\sigma(\mathbf{E})|\sigma(\mathbf{E_t}))\\
 &~+\mathbf{v}_1^T(\mathbf{y}-\mathbf{Dx-e})+\frac{\rho_1}{2}\|\mathbf{y}-\mathbf{Dx-e}\|_2^2.
\end{aligned}
\end{equation}
To calculate $\mathbf{e}_{t+1}$, we consider a two-step fast approximation. In Step 1, we first solve the following problem:
\begin{equation}
\begin{aligned}
&\hat{\mathbf{e}}=\underset{\mathbf{e}}{\argmin}~~f_{\text{GC}}(\mathbf{e}|\mathbf{e}_t)+\mathbf{v}_1^T(\mathbf{y}-\mathbf{Dx-e})\\
&~~~~~~~~~~~~~~~~~~~~+\frac{\rho_1}{2}\|\mathbf{y}-\mathbf{Dx-e}\|_2^2.
\end{aligned}
\end{equation}
According to equation (22), the surrogate function for the weight learning function is reformulated as

\begin{equation}
\begin{aligned}
 f_{\text{GC}}(\mathbf{e}|\mathbf{e}_t)&=f_{\text{GC}}'(\mathbf{e}|\mathbf{e}_t)\mathbf{e}=[\gamma_{\alpha,\beta}(1-\text{exp}(-\lambda|\mathbf{e}_t|^{\alpha}))]'\mathbf{e}\\
& = [\gamma_{\alpha,\beta}(1-\text{exp}(-\lambda|\mathbf{e}_t^2|^{\frac{\alpha}{2}}))]'\mathbf{e}^2\\
&=\frac{1}{2}\gamma_{\alpha,\beta}\lambda\alpha\text{exp}(-\lambda|\mathbf{e}_t^2|^{\frac{\alpha}{2}})|\mathbf{e}_t|^{\frac{\alpha}{2}-1}\mathbf{e}^2 \propto \|\sqrt{\mathbf{w}}\odot \mathbf{e}\|_2^2,
\end{aligned}
\end{equation}
where $\mathbf{w}=\text{exp}(-\lambda\|\mathbf{e}_t^2\|^{\frac{\alpha}{2}})\|\mathbf{e}_t\|^{\frac{\alpha}{2}-1}$ is actually a mask for the representation error. During optimization, larger weights will be allocated to the normal representation error while smaller weights are allocated to the outlier/occlusion, which is essential for outlier detection. $\odot$ denotes the Hadamard product. Thus equation (27) can be rewritten as
\begin{equation}
\begin{aligned}
&\hat{\mathbf{e}}=\underset{\mathbf{e}}{\argmin}~~\|\sqrt{\mathbf{w}}\odot \mathbf{e}\|_2^2+\mathbf{v}_1^T(\mathbf{y}-\mathbf{Dx-e})
+\frac{\rho_1}{2}\|\mathbf{y}-\mathbf{Dx-e}\|_2^2\\
&~~=\underset{\mathbf{e}}{\argmin}\left\|\begin{bmatrix}\sqrt{\frac{\rho_1}{2}}(\mathbf{y-Dx}+\frac{\mathbf{v}_1}{\rho_1})\\\mathbf{0}\end{bmatrix}-\begin{bmatrix}\sqrt{\frac{\rho_1}{2}}\\-\sqrt{\mathbf{w}}\end{bmatrix}\mathbf{e}\right\|_2^2.
\end{aligned}
\end{equation}
Obviously, equation (29) has a closed-form solution, i.e.,
\begin{equation}
\hat{\mathbf{e}}=(\mathbf{y-Dx}+\frac{\mathbf{v}_1}{\rho_1})/(\mathbf{I}+\frac{2\mathbf{w}}{\rho_1}).
\end{equation}
 With the weight obtained in equation (28), the representation error has been re-weighted to enhance the useful image content and suppress the outlier information. To well constraint the structure of the residual error, we then solve the low-rank approximation problem in Step 2 as follows.

 \begin{equation}
 \begin{aligned}
 \mathbf{E}_{t+1}&\,=\underset{\mathbf{e}}{\argmin}\frac{1}{2}\|\mathbf{E-\mathbf{\hat{E}}}\|_F^2+\lambda_1f_{\text{GC}}(\sigma(\mathbf{E})|\sigma(\mathbf{E}_t))\\
 & =\underset{\mathbf{e}}{\argmin}\frac{1}{2}\|\mathbf{E-\mathbf{\hat{E}}}\|_F^2+\lambda_1\sum_{i=1}^M f'_{\text{GC}}(\sigma_i(\mathbf{E}))\sigma_i\\
 &=\underset{\mathbf{e}}{\argmin}\frac{1}{2}\|\mathbf{E-\mathbf{\hat{E}}}\|_F^2+\lambda_1\|\mathbf{E}\|_{\text{GC}^*}
 \end{aligned}
 \end{equation}
 where $\|\mathbf{E}\|_{\text{GC}^*}$ is the proposed robust low-rank approximation. $\mathbf{E}$ can be updated by $\mathbf{E}=\mathbf{U}\mathbf{\Sigma}\mathbf{V}^T$, where $\mathbf{\Sigma}=\text{diag}(a_1,\cdots, a_m)$ with $a_i=\text{max}(\sigma_i-w_i\lambda_1,0)$. $\mathbf{U}\text{diag}(\sigma_1,\cdots, \sigma_m)\mathbf{V}^T$ is the singular value decomposition (SVD) of $\hat{\mathbf{E}}$, and $w_i=g'(\sigma_i(\mathbf{E}))$.
 Then the optimal $\mathbf{e}_{t+1}$ is obtained by vectorizing $\mathbf{E}$.
 \subsubsection{Updating $\mathbf{h}_{t+1}$}
In this paper, we use the $l_2$-norm to regularize the coefficient $\mathbf{x}$ (or $\mathbf{h}$). We update $\mathbf{h}_{t+1}$ by solving
\begin{equation}
\begin{aligned}
&\mathbf{h}_{t+1}=\underset{\mathbf{h}}{\argmin}~\lambda_2 \|\mathbf{h}\|_2^2+\frac{\rho_2}{2}\|\mathbf{x}_t-\mathbf{h}\|_2^2+\mathbf{v}_2^T(\mathbf{x}_t-\mathbf{h})\\
&~~~~~~~~ =(\mathbf{x}_t+\frac{\mathbf{v}_2}{\rho_2})_+,
\end{aligned}
\end{equation}
where $(\cdot)_+$ is a threshold function used to adjust the negative
values of its arguments to zeros while keeping the positive ones.
\subsubsection{Updating $\mathbf{x}_{t+1}$}
The update of coefficient is obtained by solving
\begin{equation}
\begin{aligned}
&\!\!\!\!\!\mathbf{x}_{t+1}\!=\underset{\mathbf{x}}{\argmin} \frac{\rho_1}{2}\|\mathbf{y}-\mathbf{D}\mathbf{x}\!-\mathbf{e}_{t+1}\|_2^2+\mathbf{v}_{1,t}^T(\mathbf{y}-\!\mathbf{D}\mathbf{x}\!-\mathbf{e}_{t+1})\\
&~~~~~~~~~~~+\mathbf{v}_{2,t}^T(\mathbf{x}-\mathbf{h}_{t+1})+\frac{\rho_2}{2}\|\mathbf{x}-\mathbf{h}_{t+1}\|_2^2.
 \end{aligned}
\end{equation}
The optimal $\mathbf{x}_{t+1}$ can be obtained by solving the following problem:
\begin{equation}
\begin{aligned}
\underset{\mathbf{x}}{\min}
\left\|\begin{bmatrix} \sqrt{\frac{\rho_1}{2}}(\mathbf{y}-\mathbf{e}_{t+1}+\frac{\mathbf{v}_{1,t}}{\rho_1})\\\sqrt{\frac{\rho_2}{2}}(\mathbf{h}_{t+1}-\frac{\mathbf{v}_{2,t}}{\rho_2})\end{bmatrix}-\begin{bmatrix}\sqrt{\frac{\rho_1}{2}}\mathbf{D}\\\sqrt{\frac{\rho_2}{2}}\mathbf{I}\end{bmatrix}\mathbf{x}\right\|_2^2.
\end{aligned}
\end{equation}
Thus, the optimal $\mathbf{x}_{t+1}$ is given by
\begin{equation}
\begin{aligned}
\mathbf{x}_{t+1}=\frac{\rho_1\mathbf{D}^T(\mathbf{y}-\mathbf{e}_{t+1}+\frac{\mathbf{v}_{1,t}}{\rho_1})
+\rho_2(\mathbf{h}_{t+1}-\frac{\mathbf{v}_{2,t}}{\rho_2})}{\rho_1\mathbf{D}^T\mathbf{D}+\rho_2\mathbf{I}}.
 \end{aligned}
\end{equation}
The proposed UWLLA algorithm is summarized in Algorithm 1.
\begin{algorithm}
\caption{The UWLLA  Algorithm}
\begin{algorithmic}[1]
\Require
   Given a test image $\mathbf{y}\in R^{m}$, and a set of training images $\mathbf{D}=[\mathbf{d}_1,\mathbf{d}_2,\cdots, \mathbf{d}_n]\in R^{m\times n}$ with each $\mathbf{d}_i\in R^{m}$ being a training sample, $\alpha, \beta, \lambda_1, \lambda_2, \rho_1$ and $\rho_2$. Initialize $\mathbf{e}$ with $\mathbf{y}$, $\mathbf{h}= \mathbf{0}$, $\mathbf{x}=\mathbf{1}/n$, $\mathbf{v}_1= \mathbf{0}$, $\mathbf{v}_2= \mathbf{0}$.
\Ensure
 $\mathbf{x}^*,\mathbf{w}^*$.
\While {$t=1,\dots,T$}\\
\textbf{Majorization step}:\\
Generate surrogate function for the objection function (15) using (22).\\
\textbf{Minimization step}:
\State Update weights by \\ ~~~~~~~$\mathbf{w}=\text{exp}(-\lambda\|\mathbf{e}_t^2\|^{\frac{\alpha}{2}}) \|\mathbf{e}_t\|^{\frac{\alpha}{2}-1}$;
\State Update $\mathbf{e}_{t+1}$ by (26)-(31);
\State Update $\mathbf{h}_{t+1}$ by (32);
\State Update $\mathbf{x}_{t+1}$ by (33-35);
\State Update $\mathbf{v}_{1,t+1}$ by $\mathbf{v}_{1,t+1}=\mathbf{v}_{1,t}+\rho_1(\mathbf{y}-\mathbf{D}\mathbf{x}_{t+1}-\mathbf{e}_{t+1})$;
\State Update $\mathbf{v}_{2,t+1}$ by $\mathbf{v}_{2,t+1}\!\!=\!\mathbf{v}_{2,t}\!+\!\rho_2(\mathbf{x}_{t+\!1}\!-\!\mathbf{h}_{t+\!1})$;
\If {$\epsilon>1e-5$}
\State repeat;
\Else
\State $t\leftarrow t+1$; Break;
\EndIf
\EndWhile
\end{algorithmic}
\end{algorithm}

\subsection{Identification}
In sparse representation-based classification, the validity of classifying samples largely depends on the reconstruction error from a specific class. Thus how to design an effective criterion to calculate the reconstruction error becomes a key issue. Let $k_i(\mathbf{x}^*)$ be the sparse coefficients corresponding to the class $i, i=1,\cdots, C$, for each class. We obtain the approximated representation as $\mathbf{D}k_i(\mathbf{x}^*)$. In this work, based on the optimal solution $\mathbf{x}^*$ and $\mathbf{w}^*$ obtained in algorithm 1, the residual between the test sample and the approximated representation for each class $i$ is defined by
\begin{equation}
\mathbf{e}_i(\mathbf{y})=\|\sqrt{\mathbf{w}^*}\odot(\mathbf{y}-\mathbf{D}k_i(\mathbf{x}^*))\|_2^2.
\end{equation}
where $k_i(\mathbf{x}^*)$ is a subvector of $\mathbf{x}^*$ corresponding to the coefficients of training samples from the $i$-th class. Then the label of the test sample $\mathbf{y}$ is defined by
\begin{equation}
\text{Identity} (\mathbf{y})=\argmin_i \{\mathbf{e}_i(\mathbf{y})\}_{i=1}^C.
\end{equation}

 \section{Computational Complexity and Convergence Analysis}
\subsection{Computational Complexity}
 Suppose $\mathbf{y}\in R^m$ is a testing image vector and $\mathbf{Y}\in R^{m_1\times m_2}(m_1\leq m_2)$ is its matrix form, and the training set is $\mathbf{D}\in R^{m\times n}$. The computational complexity for Step 6 in Algorithm 1 is $O(n)$, for Step 7 is $O(mn+m_1m_2^2)$ which is determined by the matrix multiplication $\mathbf{Dx}$, and the SVD of matrix $\mathbf{E}$. Step 9 requires $mn$ multiplications for $\mathbf{D}^T\mathbf{y}$. Thus, the total computational complexity for Algorithm 1 is $O(T(n+mn+m_1m_2^2+mn))$, where $T$ is the number of iterations.
 \subsection{Convergence Analysis}
Since the original objective  function in equation (15) is non-convex, it is difficult to determine the existence and unicity of its minimum. However, after constructing the surrogate functions for the weight learning and low-rank approximation, the reformulated objective function (23) is convex with variables $(\mathbf{e},\mathbf{E},\mathbf{x})$. Based on equation (23), we here give a theoretical analysis for the convergence property of the proposed UWLLA algorithm to show that any accumulation point of the iteration sequence constructed by the proposed model (23) is a stationary point that satisfies the Karush-Kuhn-Tucker (KKT) condition~\cite{lu2018structurally}\cite{lu2018low}.

When the UWLLA algorithm converges to a stationary point, the KKT conditions of the objective function in equation (23) are given as follows:
\begin{equation}
\begin{aligned}
& \mathbf{y}-\mathbf{Dx}-\mathbf{e}=0\\
& \mathbf{x}-\mathbf{h}=0\\
& (2w+\rho_1)\mathbf{e}-\mathbf{v}_{1,t}+\rho_1(\mathbf{y}-\mathbf{Dx})=0\\
& \mathbf{h}-(\mathbf{x}_t+\frac{\mathbf{v}_2}{\rho_2})_+=0\\
& (\rho_1\mathbf{D}^T\mathbf{D}+\rho_2)\mathbf{x}-\rho_1\mathbf{D}^T\mathbf{y}+\rho_1\mathbf{D}^T\mathbf{e}_{t+1}-\mathbf{D}^T\mathbf{v}_{1,t}
 +\mathbf{v}_{2,t}-\rho_2\mathbf{h}_{t+1}=0.
\end{aligned}
\end{equation}
Since the procedure for solving the low-rank matrix $\mathbf{E}$ is not involved in the Lagrange Multipliers, the KKT condition for it is not considered here. We then prove that the proposed algorithm converges to a point that satisfies the KKT conditions.

\textbf{Theorem 1}: Let $\mathcal{S}\triangleq(\mathbf{e},\mathbf{x},\mathbf{h},\mathbf{v}_1,\mathbf{v}_2,\rho_1, \rho_2)$ and $\{\mathcal{S}_i\}_i^{\infty}$ be the sequences generated by the proposed UWLLA algorithm. Assume that $\{\mathcal{S}_i\}_i^{\infty}$ is bounded and $\lim_{i\rightarrow \infty}\{\mathcal{S}_{i+1}-\mathcal{S}_i\}=0$, then any accumulation point of $\{\mathcal{S}_i\}_i^{\infty}$ satisfies the KKT conditions. Specifically, whenever $\{\mathcal{S}_i\}_i^{\infty}$ converges, it converges to a KKT point.

\textbf{Proof}: According to the algorithm 1, we can obtain the Lagrange multipliers $\mathbf{v}_1$ and $\mathbf{v}_2$ as follows:
\begin{equation}
\begin{aligned}
&\mathbf{v}_{1,t+1}=\mathbf{v}_{1,t}+\rho_1(\mathbf{y}-\mathbf{D}\mathbf{x}^{t+1}-\mathbf{e}^{t+1})\\
&\mathbf{v}_{2,t+1}=\mathbf{v}_{2,t}+\rho_2(\mathbf{x}^{t+1}-\mathbf{h}^{t+1}),
\end{aligned}
\end{equation}
where $\mathbf{v}_{j,t+1}$ for $j=1,2$ is the next point of $\mathbf{v}_{j,t}$ in a sequence $\{\mathbf{v}_i^j\}_{i=1}^{\infty}$. If sequences $\{\mathbf{v}_1^i\}_{i=1}^{\infty}$ and $\{\mathbf{v}_2^i\}_{i=1}^{\infty}$ converges to a stationary point, i.e., $(\mathbf{v}_{1,t+1}-\mathbf{v}_{1,t})\rightarrow 0$ and $(\mathbf{v}_{2,t+1}-\mathbf{v}_{2,t})\rightarrow 0$, then $(\mathbf{y}-\mathbf{D}\mathbf{x}^{t+1}-\mathbf{e}^{t+1})\rightarrow 0$ and $(\mathbf{x}_{t+1}-\mathbf{h}_{t+1})\rightarrow 0$. Thus, the KKT conditions related to variables $\mathbf{v}_1$ and $\mathbf{v}_2$ are satisfied.

We then derive the third KKT condition according to the proposed algorithm as follows:
\begin{equation}
\begin{aligned}
(2w+\rho_1)(\mathbf{e}_{t+1}-\mathbf{e}) = \mathbf{v}_{1,t}-\rho_1(\mathbf{y-\mathbf{Dx}})-(2w+\rho_1)\mathbf{e}.
\end{aligned}
\end{equation}
When $(\mathbf{e}_{t+1}-\mathbf{e})\rightarrow 0$, we have $\mathbf{v}_{1,t}-\rho_1(\mathbf{y}-\mathbf{Dx})-(2w+\rho_1)\mathbf{e} \rightarrow 0$.

\noindent Then the fourth KKT condition can be derived as follows:
\begin{equation}
\begin{aligned}
\mathbf{h}_{t+1}-\mathbf{h} = (\mathbf{x}_t+\frac{\mathbf{v}_2}{\rho_2})_+ -\mathbf{h}.
\end{aligned}
\end{equation}
When $\mathbf{h}^{t+1}\rightarrow \mathbf{h}$, we then have $(\mathbf{x}_t+\frac{\mathbf{v}_2}{\rho_2})_+ -\mathbf{h} \rightarrow 0$.

The fifth KKT condition is
\begin{equation}
\begin{aligned}
&(\rho_1\mathbf{D}^T\mathbf{D}+\rho_2)(\mathbf{x}^{t+1}-\mathbf{x}) = \rho_1\mathbf{D}^T\mathbf{y}-\rho_1\mathbf{D}^T\mathbf{e}_{t+1}+\mathbf{D}^T\mathbf{v}_{1,t}\\
&-\mathbf{v}_{2,t}+\rho_2\mathbf{h}_{t+1}-(\rho_1\mathbf{D}^T\mathbf{D}+\rho_2)\mathbf{x},
\end{aligned}
\end{equation}
from which we can see that when $\mathbf{x}^{t+1}-\mathbf{x} \rightarrow 0$, we have $\rho_1\mathbf{D}^T\mathbf{e}_{t+1}+\mathbf{D}^T\mathbf{v}_{1,t}-\mathbf{v}_{2,t}+\rho_2\mathbf{h}_{t+1}-(\rho_1\mathbf{D}^T\mathbf{D}+\rho_2)\mathbf{x}\rightarrow 0$. Based on the above analysis,
$\lim_{i\rightarrow \infty} \{\mathcal{S}_{i+1}-\mathcal{S}_i\}=0$ indicates that both sides of the equations (39-42) approach zeros as $i\rightarrow \infty$. Therefore, in an asymptotic sense, the sequence $\{\mathcal{S}_i\}_{i}^{\infty}$ satisfies the KKT condition of (23).
\section{Experimental results}
 \subsection{Databases and Parameter Settings}
 To verify the effectiveness of the proposed method, we carry out  experiments on three publicly available face databases, including Extended Yale B (ExYaleB)~\cite{georg2001few}, AR~\cite{AR1998}, and aligned Labeled Face in the Wild (LFW-a)~\cite{wolf2009similarity}. The ExYaleB database are captured under 576 viewing conditions (9 poses $\times$ 64 illuminations conditions), and can be regarded as a databases that can cause complex error distributions. We will apply additional random pixel noises and block occlusions to the original images to test the efficacy of the proposed method. The AR database consists of face images with real disguise including sunglasses and scarf occlusions. Random pixel noises are added to images to simulate complex error distributions. The LFW-a database consists of face images captured in an unconstrained environment with varying poses, expression, lighting, and is used to test the performance of the proposed method in real-world environment.

 Since the proposed method is a regression-based model, we test and compare it with 8 recently published regression-based face recognition approaches, including RRC-L1 and RRC-L2~\cite{yang2012regularized}, HQ-A and HQ-M~\cite{he2013half}, F-LR-IRNNLS~\cite{iliadis2017robust}, IRGSC~\cite{zheng2017iterative}, NMR~\cite{yang2016nuclear}, and LUM~\cite{dong2019low}.
 $\alpha$ and $\beta$ are two important parameters in the proposed model, the former models the shape of the error distribution, and the latter is the kernel width. Let $\alpha_1$ and $\beta_1$ be the parameters defined for the weight learning, and $\alpha_2$ and $\beta_2$ for the rank approximation. In this paper, $1<\alpha_1<2$ is used for weight learning for all the experiments. We fixed $\alpha_2=1$ and $\beta_2=0.7$ for a better low-rank approximation for all the experiments. For the proposed ADMM optimization algorithm, we use $\rho_1=1$, $\rho_2=0.1$, $\lambda_1=0.01$, and $\lambda_2=1$.

\subsection{Experiments on the ExYaleB database}
In this experiment, images from the ExYaleB face database are used to test the robustness of the proposed algorithm. First, all the images are resized to $96\times 84$ pixels. We adopt two experimental settings for testing data, one is the data with different percentages of occlusions, the other is the data with different levels of occlusion-pixel mixed corruptions. For both experiments, we choose all the images in subsets 1 and 2 for training, and subset 3 for testing. Thus, the total number of images for training and testing are 719 and 455, respectively.
\begin{figure}[H]
\centering
    \hspace{0cm}
    \vspace{0cm}
   \subfigure[]{\includegraphics[width=0.13\columnwidth]{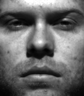}}
   \hspace{0.5cm}
   \subfigure[]{\includegraphics[width=0.13\columnwidth]{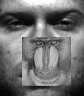}}
   \hspace{0.5cm}
   \subfigure[]{\includegraphics[width=0.13\columnwidth]{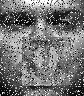}}
    \caption{Face images of the ExYaleB database with different types of corruptions. (a) A clean face image. (b) with $30\%$ percent of block occlusion. (c) with $30\%$ occlusion-pixel mixed corruption. Here $30\%$ mixed noise means $30\%$ random pixel corruption plus $30\%$ block occlusions.}
\end{figure}


\begin{table*}  
\caption {The recognition accuracies ($\%$) and running time (second (s)/per sample) of all the methods on the ExYaleB face database with $60\%$ block occlusion ($60\%$ Occ) and $60\%$ occlusion-pixel mixed corruption ($60\%$ Mix).}
\vspace{0.3cm}
 \centering
 \scalebox{0.75}{
 \begin{tabular}{|p{55pt}<{\centering}| p{35pt}<{\centering} | p{35pt}<{\centering}| p{35pt}<{\centering} | p{33.5pt}<{\centering}|p{60pt}<{\centering}|p{33pt}<{\centering}|p{30pt}<{\centering}|p{30pt}<{\centering}|p{43pt}<{\centering}|}
   \hline
   \multirow{2}{*}{Corruptions}&
   \multicolumn{9}{c|}{Methods}\\
   \cline{2-10}
   & RRC-L1 & RRC-L2 & HQ-A& HQ-M& FLR-IRNNLS& IRGSC& NMR& LUM &Proposed\\
   \hline
   $60\%$ Occ   & 69.67 & 70.54 & 48.02& 68.13 & 95.82  & 66.15 &79.12 & 89.42 & \textbf{98.46} \\
   \hline
   $60\%$ Mix  &34.72  &35.60 &17.92 &32.15 &49.01 &27.91 &8.64 & 61.40& \textbf{64.75}  \\
   \hline
   \hline
   Time (s)  &2.72  &1.30 &9.15 &10.11 &1.66 &2.97 &1.03 & 28.57& 8.01 \\
   \hline
 \end{tabular}}
 \end{table*}

 \begin{figure}[H]
\centering
    \hspace{0cm}
    \vspace{0cm}
   \includegraphics[width=0.9\columnwidth]{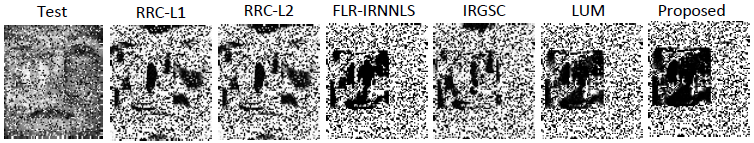}
    \caption{Weight images of different methods for an image from the ExYaleB database with $30\%$ mixed corruptions.}
\end{figure}

 In the first experiment, we evaluate the proposed method on the dataset with occlusion percentage varying from $20\%$ to $70\%$. To simulate occlusions, we randomly selected local region in each testing image and replace this area with an unrelated image. In this experiment, we use the baboon image, as used in \cite{iliadis2017robust,xie2017robust}, for occlusion. To simulate a specific percentage of occlusion for a testing image $\mathbf{Y}\in R^{m_1\times m_2}$, we resize the baboon image to $z\times z$, where $z=\sqrt{m_1\times m_2 \times x\%}$ and replace the local region in the testing image. One example of the occluded testing image is shown in Figure 2(b). We plot the weight images for the proposed method and for the methods which also focusing on learning weights for the corrupted images, as shown in Figure 3. The weight images show that the proposed method can detect the whole structure of occlusion in the presence of the heavy pixel noise corruptions, which means that it
 has superior ability in characterizing the structure of the occlusion under complex corruption conditions.  The recognition accuracies of all methods on the data with $60\%$ occlusion are shown in Table 1 where the proposed algorithm obtained the highest accuracy $98.46\%$. The recognition rates from the proposed method and all the benchmarks under different percentage
of occlusions are shown in Figure 4(a), which show that our method recognizes the testing images with nearly $100\%$ accuracy when occlusion percentage is not larger than $50\%$, and still achieves the highest accuracy when occlusion percentage is larger than $50\%$. Especially, the accuracy of the proposed method are nearly $3\%$ and $30\%$ higher than that of the second best performing method under $60\%$ and $70\%$ occlusions, respectively.
 \begin{figure}[H]
\centering
    \hspace{-0.3cm}
    \vspace{0cm}
   \subfigure[]{\includegraphics[width=0.48\columnwidth]{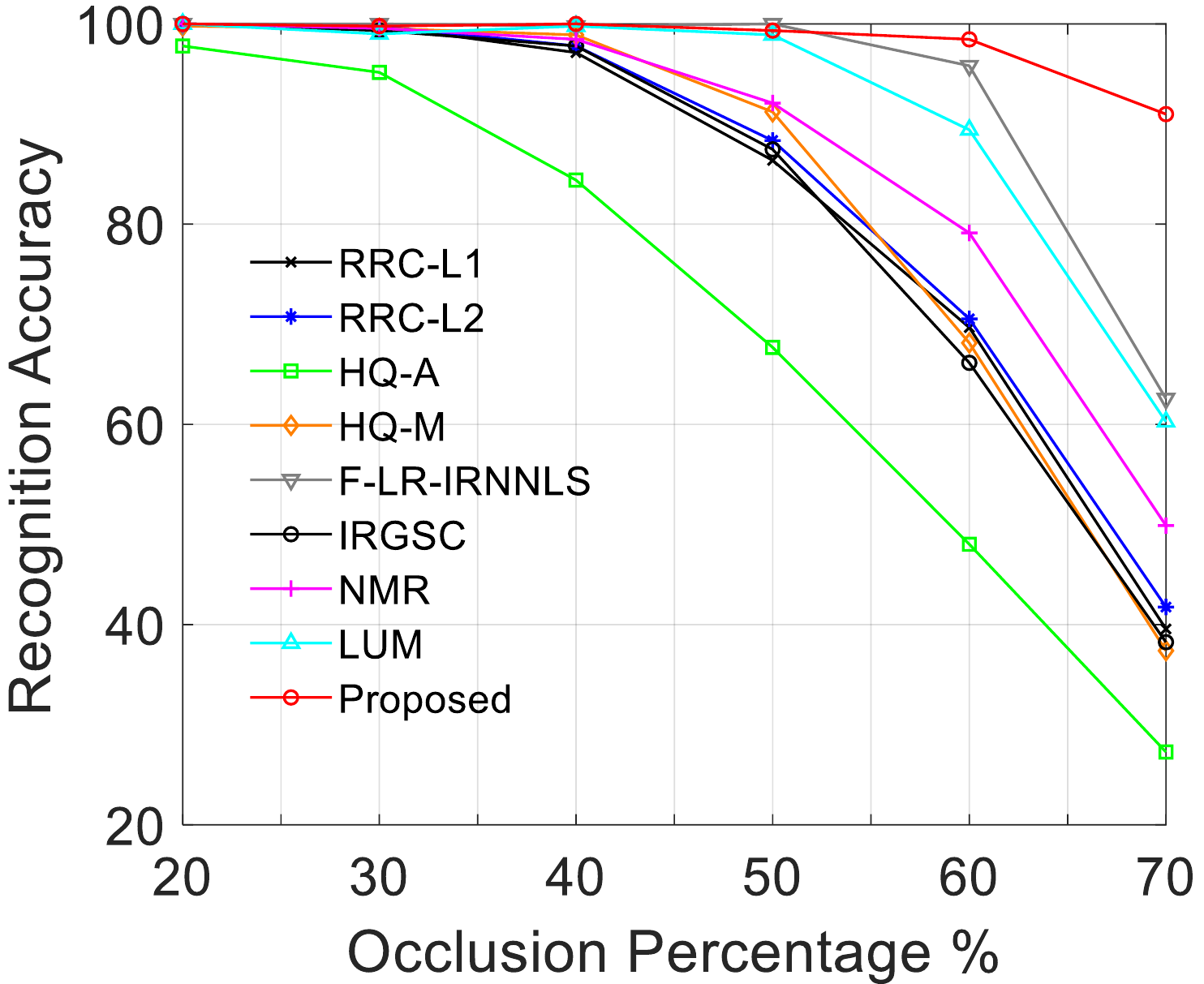}}
   \subfigure[]{\includegraphics[width=0.48\columnwidth]{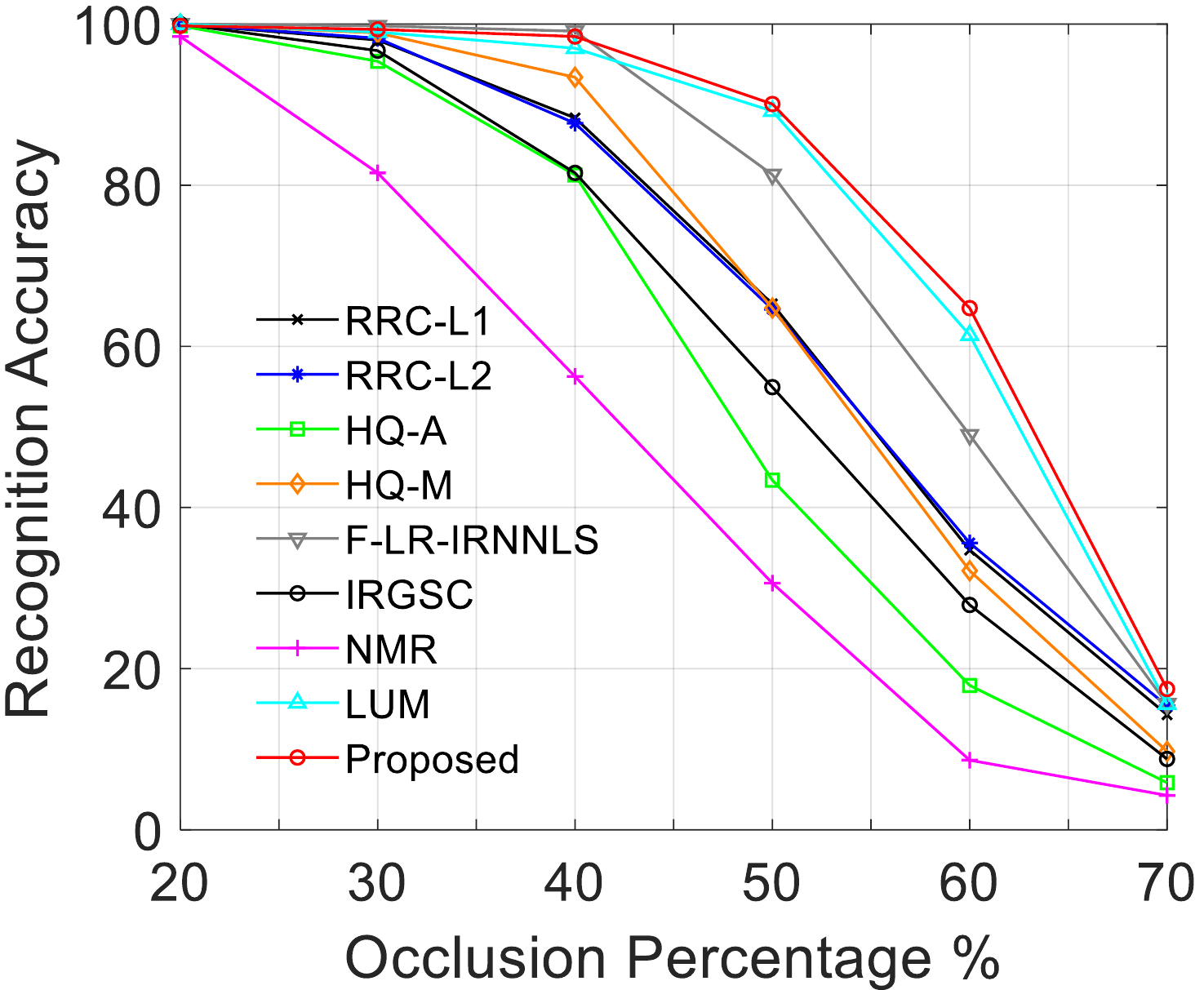}}
    \caption{Recognition accuracies with different types of corruptions on the ExYaleB database. (a) with different percentage of occlusions. (b) with mixed corruptions.}
\end{figure}
To evaluate the performance of the proposed method under more challenging conditions, we test it on the dataset with different levels of mixed pixel corruptions and block occlusions. The corruption level varies from $20\%$ to $70\%$. To simulated the mixed corruptions, we randomly select a certain percent of pixels and replace them with random values. An example image with $30\%$ percent of mixed corruptions is shown in Figure 2(c). The recognition accuracies of all the methods with $60\%$ mixed corruptions are shown in Table 1 where the proposed method obtains the best accuracy. The accuracies of all the methods with varying levels of corruptions are shown in Figure 4(b) which indicates that the proposed algorithm can tolerate the mixed corruptions very well, and is superior to other benchmarks. We also compares the running time of each algorithm for recognizing one sample in Table 1, which shows that the computational complexity of our algorithm is acceptable.

\subsection{Experiments on the AR database}
To evaluate the robustness of the proposed method in recognizing face images with real disguise, we test it on the AR database which contains face images with sunglasses and scarf occlusions. In AR database, there are two sessions of facial images from 100 subjects (50 male and 50 female). In each session, there are 2 natural unoccluded face images, 3 face images with scarf disguise and 3 with sunglasses. We first test all the algorithms on the images with real disguise, and then on face images with combined corruptions (real disguise and random noise). Some example images from one individual are shown in Figure 5.

 \begin{figure}[H]
\centering
    \hspace{0cm}
    \vspace{0cm}
   \subfigure[]{\includegraphics[width=0.12\columnwidth]{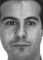}}
   \hspace{0.1cm}
   \subfigure[]{\includegraphics[width=0.12\columnwidth]{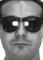}}
   \hspace{0.1cm}
   \subfigure[]{\includegraphics[width=0.12\columnwidth]{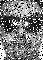}}
   \hspace{0.1cm}
   \subfigure[]{\includegraphics[width=0.12\columnwidth]{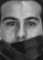}}
   \hspace{0.1cm}
   \subfigure[]{\includegraphics[width=0.12\columnwidth]{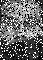}}
    \caption{Face images of the AR database with different corruptions. (a) A clean face image. (b) with sunglass occlusion. (c) with sunglass-pixel corruption. (d) with scarf occlusion. (e) with scarf-pixel corruption.}
\end{figure}

\begin{figure}[H]
\centering
    \hspace{0cm}
    \vspace{0cm}
   \includegraphics[width=0.8\columnwidth]{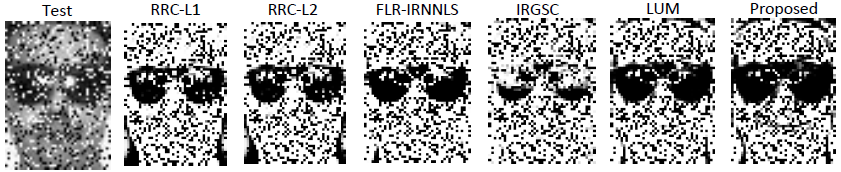}
    \caption{Weight images of different methods for an image from the AR database with $30\%$ mixed corruptions.}
\end{figure}

We first test the proposed method on session 1 and 2 separately. For each session, we select the only natural image from each individual as the training image, and 3 images with sunglasses and 3 images with scarf for testing. Then, we test the proposed method on the data from both sessions, where two natural images of each person are selected as training images, and 6 images with sunglasses and 6 images with scarf are used for testing. The weight images learned by the proposed method and five benchmarks for a sunglasses-pixel corrupted image are shown in Figure 6. As can be seen from these figures, the proposed method can learn more accurate weights for the corruptions than other methods because it finds out the whole structure of the sunglasses while other methods only detect part of them.
The recognition rates from the proposed method and all the benchmarks are shown in Table 2, which demonstrates that our method outperforms all the benchmarks in terms of single session and both session testing under two different types of occlusions.
For the sunglasses case, the proposed method achieves around $93\%$ for all sessions, which is much higher than other approaches. In particular, in session 1, the recognition accuracy of the proposed method is $5.21\%$ higher than the second highest one. For experiments with scarf occlusions, the performance of the proposed method is $1.3\%-3.8\%$ higher than the method with the second highest accuracy.

\begin{table*}
 \caption {The recognition accuracies ($\%$) of all the algorithms on the AR face database with different occlusions.}
 \vspace{0.3cm}
 \centering
 \scalebox{0.75}{
 \begin{tabular}{|p{60pt} <{\centering}| p{37pt}<{\centering} | p{32pt}<{\centering}| p{30pt}<{\centering} | p{30pt}<{\centering} |p{32pt}<{\centering}|p{40pt}<{\centering}|p{32pt}<{\centering}|p{23pt}<{\centering}|p{23pt}<{\centering}|p{40pt}<{\centering}|}
    \hline
    \multicolumn{2}{|l|}{ \multirow{2}{*}{Evaluation Types}}&
   \multicolumn{9}{c|}{Methods }\\
   \cline{3-11}
   \multicolumn{2}{|l|}{ }  & RRC-L1 & RRC-L2 & HQ-A& HQ-M&FLR-IRNNLS& IRGSC& NMR& LUM&Proposed \\
   \hline
   \multirow{2}*{Session 1}& sunglass  & 75.33 & 77.00 & 68.00 & 72.67&83.33& 73.33&75.67& 88.21&\textbf{93.00}\\
    & scarf  & 61.66 & 64.33 & 30.54 & 35.23&55.67& 54.33&60.74& 65.03&\textbf{66.33} \\
   \hline
   \multirow{2}*{Session 2}& sunglass  & 82.66 & 83.66 & 70.13& 72.48&86.24& 77.67&80.47& 91.15&\textbf{93.67} \\
    & scarf  & 60.66 & 61.66 & 25.33 & 30.00&48.00& 54.33&54.00& 61.18&\textbf{63.00} \\
   \hline
   \multirow{2}*{Both sessions} & sunglass  & 83.00 & 83.16& 70.23 & 72.58&85.79& 77.33&72.53& 91.87&\textbf{94.67}\\
    & scarf  & 67.33 & 69.33 & 29.67 & 34.17&57.33& 62.83&64.00& 67.50&\textbf{71.33}\\
    \hline
 \end{tabular}}
 \end{table*}
 To test the proposed method in more complicated cases, we further conduct experiments on the face images with different levels of sunglasses-random pixel mixed and scarf-random pixel mixed corruptions. With the level of the mixed noise varying from $20\%$ to $70\%$, the performance of all the methods are shown in Figure 7 in which the accuracy curves show that the proposed method clearly outperforms the benchmarks. Furthermore, the recognition accuracy curves of the proposed method and LUM decrease slightly when the percentage of mixed noises increases from $20\%$ to $60\%$, while the curves for other methods show bigger decrease in this range.

 \begin{figure}[H]
\centering
    \hspace{-0.3cm}
    \vspace{0cm}
   \subfigure[]{\includegraphics[width=0.5\columnwidth]{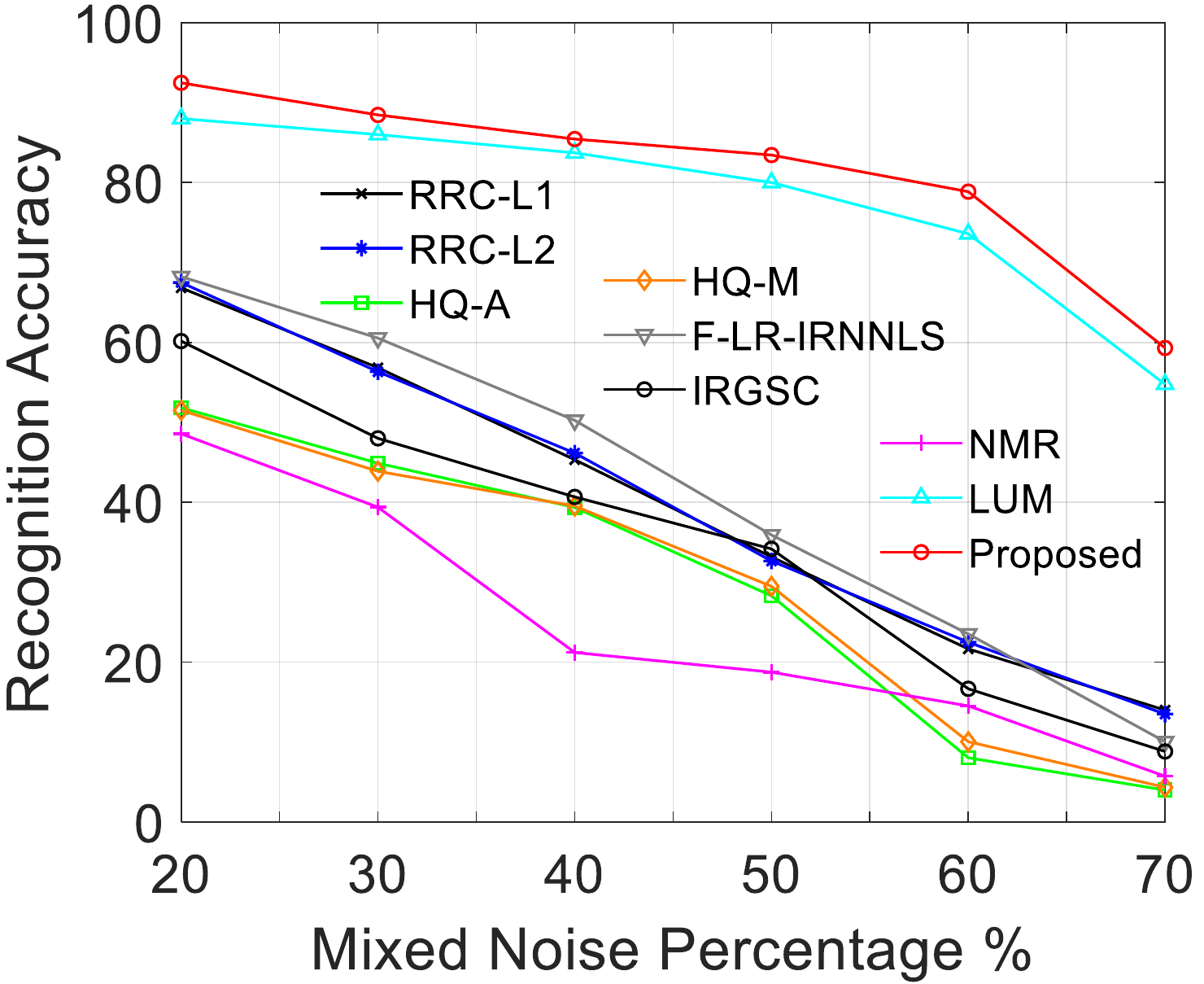}}\,
   \subfigure[]{\includegraphics[width=0.5\columnwidth]{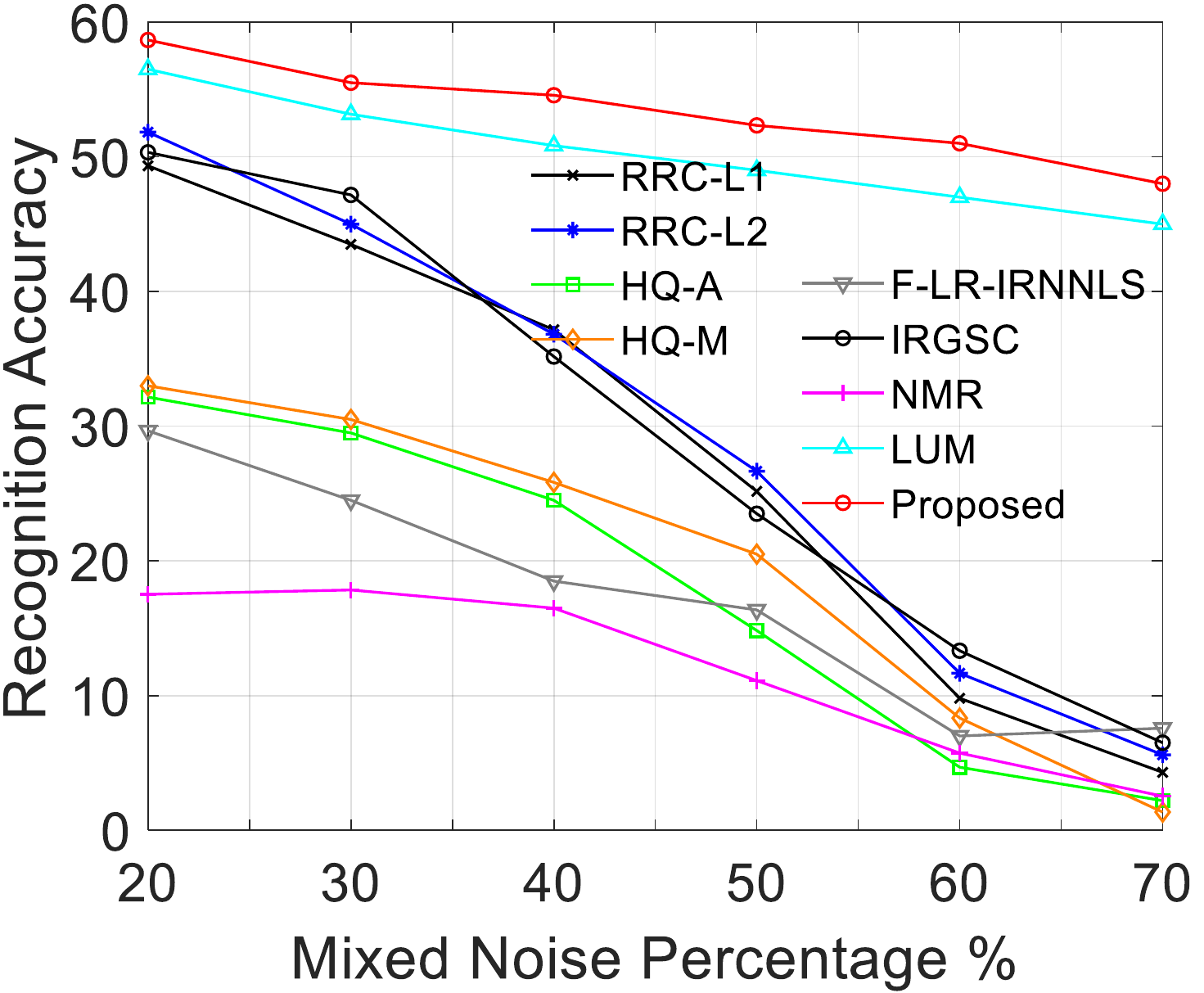}}
    \caption{Recognition accuracies under different types of corruptions on the AR database. (a) under different percentages of sunglasses-random pixel corruptions; (b) under different percentages  scarf-random pixel mixed corruptions.}
\end{figure}

\subsection{Experiments on the LFW-a database}
To evaluate the robustness of the proposed algorithm for face recognition under an unconstrained environment, we carry out experiments on the LFW database. The LFW database is a database of face photographs designed for studying the problem of unconstrained face recognition. In this paper, we use the aligned version LFW-a database for all the experiments. We use the same types of corruptions as in Experiment 1, i.e., block occlusion (baboon image) and occlusion-pixel mixed corruption, to further evaluate the robustness of the proposed method in this experiment. The clean images and their corrupted counterparts from one individual are shown in Figure 8. We select 158 subjects and each of them has at least 10 images. For each subject, we randomly select 5 samples for training and 5 samples for testing. Thus, the number of training samples and testing samples are both 790.
\begin{figure}[H]
\centering
    \hspace{0cm}
    \vspace{0cm}
   \includegraphics[width=0.7\columnwidth]{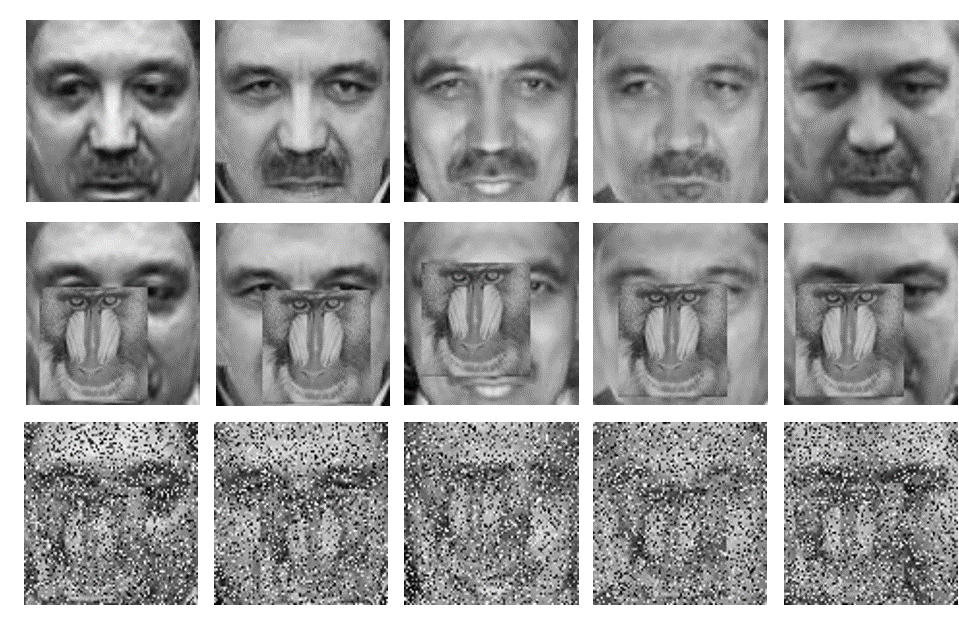}
   \vspace{-0.3cm}
    \caption{Face images of one subject in the LFW-a face database with different corruptions. 1st row: original images; 2nd row: images with block occlusions; 3rd row: images with mixed corruptions.}
\end{figure}
 \vspace{-0.5cm}
\begin{figure}[H]
\centering
    \hspace{0cm}
    \vspace{0cm}
   \includegraphics[width=1\columnwidth]{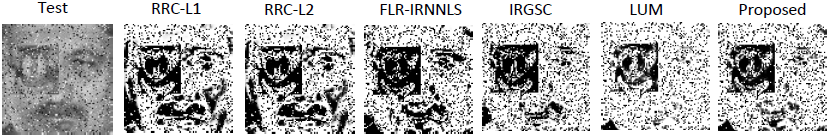}
    \caption{Weight images of different methods for an image from the LFW database with $20\%$ mixed corruptions.}
\end{figure}

First, we test all the methods on the clean images to verify the effectiveness of the proposed method in tolerating the error caused by illumination, pose, and expression changes (unconstrained conditions). Then all the methods are tested on the data with $20\%$ block occlusion ($20\%$ Occ.), $20\%$ mixed corruptions ($20\%$ Mix.), $40\%$ Occ., and $40\%$ Mix.. We plot the weight images from the proposed method and benchmarks in Figure 9 from which we can see that RRC-L1, RRC-L2, FLR-IRNNLS, and IRGSC have serious mis-weighting for corrruptions since the normal face regions are recognized as outliers. Although both the LUM method and the proposed method don't show much mis-weighting problems, the LUM method has inferior ability in detecting the edge of occlusion while the proposed method can characterize the edge better. Thus, the proposed method is less sensitive to the representation errors caused by unconstrained environment than other methods. The recognition accuracies of different methods on the clean and corrupted datasets are reported in Table 3 which shows that the proposed method can better tolerate representation errors caused by noise, occlusions and other unconstrained environment.

\begin{table}  
\caption {The recognition accuracies ($\%$) of all the methods on the clean LFW database and data with various corruptions.}
\vspace{0.3cm}
 \centering
 \scalebox{0.75}{
 \begin{tabular}{|p{80pt}<{\centering}| p{50pt}<{\centering}| p{50pt}<{\centering}| p{50pt}<{\centering} | p{50pt}<{\centering}|p{50pt}<{\centering}|}
   \hline
   \multirow{2}{*}{Methods}&
   \multicolumn{5}{c|}{Different types /percentages of corruptions}\\
   \cline{2-6}
  &Clean & $20\%$ Occ. & $20\%$ Mix. & $40\%$ Occ. & $40\%$ Mix.\\
   \hline
   RRC-L1 & 68.98 &65.06  &63.29 &47.34 &25.56  \\
   \hline
   RRC-L2 & 69.36  &60.12  &57.08 &40.63 &19.62  \\
   \hline
   HQ-A & 51.27  &51.33 &57.09 &35.28 &30.33  \\
   \hline
   HQ-M & 58.30  &58.63  &58.94 &43.91 &32.99  \\
   \hline
   FLR-IRNNLS & 71.65  &63.37  &62.74 &51.02 &37.56  \\
   \hline
   IRGSC& 73.42  & 65.33 &50.51 &37.09 &13.04  \\
   \hline
   NMR & 72.62  &54.63  &41.83 &32.23 &14.45  \\
   \hline
   LUM & 73.04  & 65.21 & 63.75 & 51.05& 42.17 \\
   \hline
   Proposed & \textbf{74.18}  &\textbf{66.78}  &\textbf{65.19} &\textbf{52.03} &\textbf{46.70}  \\
   \hline
 \end{tabular}}
 \end{table}

 \subsection{Discussion on the selection of $\alpha$ and $\beta$ }
 According to above experimental results,  we empirically choose values of $\alpha_1$ and $\beta_1$ from $\alpha_1\in[1,2]$ and $\beta_1\in [0.05,0.15]$ for the robust weight learning. When the errors are disturbed by larger outliers which will cause heavy-tailed noise, then a lower-order statistical measure (smaller $\alpha_1$) for the error is usually more robust. In this paper, we use $\alpha_1=1.7$ for all the experiments to handle different types of corruptions. Choosing the value of $\beta_1$ is also important for the proposed model. A smaller $\beta_1$ leads to a thinner distribution, while larger $\beta_1$ leads to a fatter distribution. The error with a thinner distribution is usually caused by a simple corruption, e.g., corruption from different occlusions. A fatter distribution is caused by more complicated corruptions, e.g., occlusion-pixel mixed corruptions. Thus, for all the experiments, we use a smaller $\beta_1=0.07$ for the experiments with occlusions, and a larger $\beta_1=0.11$ for the experiments with occlusion-pixel mixed corruptions. Different from matching the error distribution in weight learning, the GC function for low-rank approximation tries to give more emphasis on the larger singular values and thus can maintain the low-rank structure of the error image. In all the experiments, $\alpha_2=1$ and $\beta_2=0.7$ can provide a good low-rank approximation.

 \section{Conclusions}
 In this paper, we overcome the vulnerability of existing regression-based face recognition methods in dealing with complex error distributions, and investigate the relationship between the generalized correntropy and its ability in complex error modelling. With the analysis and discovery, we developed a unified sparse weight learning and low-rank regression model which incoporates the generalized correntropy to the matching of complex representation error distribution. The sparse weight learning procedure can discriminatingly weight errors caused by random noises, while the low-rank regularization can provide a more accurate approximation for structured errors. Therefore, the learned features for the classification are more robust to various corruptions. Moreover, different from existing algorithms in handling both problems with multiple functions, we unified the two targets in one function, which provides a deep investigation of the relationship between sparse weight learning and low-rank approximation. The experimental results consistently demonstrate that the proposed method outperforms state-of-the-art methods in handling complex pixel corruptions as well as the block occlusions. Since the proposed method is a regression-based model, it cannot explore the deep features of the training data, thus a combination of regression model and multi-layer deep learning model will be of interest to us in future work.


\section*{Acknowledgments}
This work is supported in part by the Industrial Transformation Research Hub Grant IH180100002.


 \bibliographystyle{elsarticle-num}


\footnotesize
\begin{thebibliography}{10}
\expandafter\ifx\csname url\endcsname\relax
  \def\url#1{\texttt{#1}}\fi
\expandafter\ifx\csname urlprefix\endcsname\relax\def\urlprefix{URL }\fi
\expandafter\ifx\csname href\endcsname\relax
  \def\href#1#2{#2} \def\path#1{#1}\fi

\bibitem{wright2009robust}
J.~Wright, A.~Y. Yang, A.~Ganesh, S.~S. Sastry, Y.~Ma, Robust face recognition
  via sparse representation, IEEE Trans. Pattern Anal. Mach. Intell. 31~(2)
  (2009) 210--227.

\bibitem{deng2017face}
W.~Deng, J.~Hu, J.~Guo, Face recognition via collaborative representation: Its
  discriminant nature and superposed representation, IEEE Trans. Pattern Anal.
  Mach. Intell. 40~(10) (2017) 2513--2521.

\bibitem{huang2013supervised}
J.~Huang, F.~Nie, H.~Huang, C.~Ding, Supervised and projected sparse coding for
  image classification, in: Twenty-Seventh AAAI, 2013, pp. 438--444.

\bibitem{he2011maximum}
R.~He, W.-S. Zheng, B.-G. Hu, Maximum correntropy criterion for robust face
  recognition, IEEE Trans. Pattern Anal. Mach. Intell. 33~(8) (2011)
  1561--1576.

\bibitem{wang2016correntropy}
Y.~Wang, Y.~Y. Tang, L.~Li, Correntropy matching pursuit with application to
  robust digit and face recognition, IEEE Trans. Cybern. 47~(6) (2016)
  1354--1366.

\bibitem{chen2016generalized}
B.~Chen, L.~Xing, H.~Zhao, N.~Zheng, J.~C. Pr{\i}ncipe, Generalized correntropy
  for robust adaptive filtering, IEEE Trans. Signal Process. 64~(13) (2016)
  3376--3387.

\bibitem{naseem2012robust}
I.~Naseem, R.~Togneri, M.~Bennamoun, Robust regression for face recognition,
  Pattern Recognit. 45~(1) (2012) 104--118.

\bibitem{yang2012regularized}
M.~Yang, L.~Zhang, J.~Yang, D.~Zhang, Regularized robust coding for face
  recognition, IEEE Trans. Image Process. 22~(5) (2012) 1753--1766.

\bibitem{zheng2017iterative}
J.~Zheng, P.~Yang, S.~Chen, G.~Shen, W.~Wang, Iterative re-constrained group
  sparse face recognition with adaptive weights learning, IEEE Trans. Image
  Process. 26~(5) (2017) 2408--2423.

\bibitem{dong2019low}
J.~Dong, H.~Zheng, L.~Lian, Low-rank laplacian-uniform mixed model for robust
  face recognition, in: CVPR, 2019, pp. 11897--11906.

\bibitem{iliadis2017robust}
M.~Iliadis, H.~Wang, R.~Molina, A.~K. Katsaggelos, Robust and low-rank
  representation for fast face identification with occlusions, IEEE Trans.
  Image Process. 26~(5) (2017) 2203--2218.

\bibitem{yang2016nuclear}
J.~Yang, L.~Luo, J.~Qian, Y.~Tai, F.~Zhang, Y.~Xu, Nuclear norm based matrix
  regression with applications to face recognition with occlusion and
  illumination changes, IEEE Trans. Pattern Anal. Mach. Intell. 39~(1) (2016)
  156--171.

\bibitem{xie2017robust}
J.~Xie, J.~Yang, J.~J. Qian, Y.~Tai, H.~M. Zhang, Robust nuclear norm-based
  matrix regression with applications to robust face recognition, IEEE Trans.
  Image Process. 26~(5) (2017) 2286--2295.

\bibitem{qian2015robust}
J.~Qian, L.~Luo, J.~Yang, F.~Zhang, Z.~Lin, Robust nuclear norm regularized
  regression for face recognition with occlusion, Pattern Recognit. 48~(10)
  (2015) 3145--3159.

\bibitem{zheng2019weighted}
J.~Zheng, K.~Lou, X.~Yang, C.~Bai, J.~Tang, Weighted mixed-norm regularized
  regression for robust face identification, IEEE Trans. Neural Netw. Learn.
  Syst. (2019).

\bibitem{luo2016robust}
L.~Luo, J.~Yang, J.~Qian, Y.~Tai, G.-F. Lu, Robust image regression based on
  the extended matrix variate power exponential distribution of dependent
  noise, IEEE Trans. Neural Netw. Learn. Syst. 28~(9) (2016) 2168--2182.

\bibitem{wang2019block}
Y.~Wang, Y.~Y. Tang, L.~Li, X.~Zheng, Block sparse representation for pattern
  classification: Theory, extensions and applications, Pattern Recognit. 88
  (2019) 198--209.

\bibitem{xie2019hyperspectral}
T.~Xie, S.~Li, B.~Sun, Hyperspectral images denoising via nonconvex regularized
  low-rank and sparse matrix decomposition, IEEE Trans. Image Process. (2019).

\bibitem{liang2020correlation}
Z.~Liang, X.~Chen, L.~Zhang, J.~Liu, Y.~Zhou, Correlation classifiers based on
  data perturbation: New formulations and algorithms, Pattern Recognit. 100
  (2020) 107106.

\bibitem{schultz2018nonsmooth}
D.~Schultz, B.~Jain, Nonsmooth analysis and subgradient methods for averaging
  in dynamic time warping spaces, Pattern Recognit. 74 (2018) 340--358.

\bibitem{chopra2010total}
A.~Chopra, H.~Lian, Total variation, adaptive total variation and nonconvex
  smoothly clipped absolute deviation penalty for denoising blocky images,
  Pattern Recognit. 43~(8) (2010) 2609--2619.

\bibitem{yang2016multi}
B.~Yang, M.~Xiang, Y.~Zhang, Multi-manifold discriminant isomap for
  visualization and classification, Pattern Recognit. 55 (2016) 215--230.

\bibitem{ren2019sinusoidal}
J.~Ren, T.~Zhang, J.~Li, P.~Stoica, Sinusoidal parameter estimation from signed
  measurements via majorization--minimization based relax, IEEE Trans. Signal
  Process. 67~(8) (2019) 2173--2186.

\bibitem{marnissi2020majorize}
Y.~Marnissi, E.~Chouzenoux, A.~Benazza-Benyahia, J.-C. Pesquet,
  Majorize--minimize adapted metropolis--hastings algorithm, IEEE Trans. Signal
  Process. 68 (2020) 2356--2369.

\bibitem{fan2020min}
W.~Fan, J.~Liang, H.~C. So, G.~Lu, Min-max metric for spectrally compatible
  waveform design via log-exponential smoothing, IEEE Trans. Signal Process. 68
  (2020) 1075--1090.

\bibitem{sun2016majorization}
Y.~Sun, P.~Babu, D.~P. Palomar, Majorization-minimization algorithms in signal
  processing, communications, and machine learning, IEEE Trans. Signal Process.
  65~(3) (2016) 794--816.

\bibitem{boyd2011distributed}
S.~Boyd, N.~Parikh, E.~Chu, B.~Peleato, J.~Eckstein, et~al., Distributed
  optimization and statistical learning via the alternating direction method of
  multipliers, Foundations and Trends{\textregistered} in Machine learning
  3~(1) (2011) 1--122.

\bibitem{hu2019doubly}
M.~Hu, S.~Chen, Doubly aligned incomplete multi-view clustering, in:
  Twenty-Seventh IJCAI, 2018, pp. 2262--2268.

\bibitem{piao2019double}
X.~Piao, Y.~Hu, J.~Gao, Y.~Sun, B.~Yin, Double nuclear norm based low rank
  representation on grassmann manifolds for clustering, in: CVPR, 2019, pp.
  12075--12084.

\bibitem{lu2018structurally}
Y.~Lu, C.~Yuan, W.~Zhu, X.~Li, Structurally incoherent low-rank nonnegative
  matrix factorization for image classification, IEEE Trans. Image Process.
  27~(11) (2018) 5248--5260.

\bibitem{lu2018low}
Y.~Lu, Z.~Lai, X.~Li, W.~K. Wong, C.~Yuan, D.~Zhang, Low-rank 2-{D}
  neighborhood preserving projection for enhanced robust image representation,
  IEEE Trans. Cybern. 49~(5) (2018) 1859--1872.

\bibitem{georg2001few}
A.~Georghiades, P.~Belhumeur, D.~Kriegman, From few to many: Illumination cone
  models for face recognition under variable lighting and pose, IEEE Trans.
  Pattern Anal. Mach. Intell. 23~(6) (2001) 643--660.

\bibitem{AR1998}
A.~Martinez, R.~Benavente, The {AR} face database, Tech. rep., The Ohio State
  University (06 1998).

\bibitem{wolf2009similarity}
L.~Wolf, T.~Hassner, Y.~Taigman, Similarity scores based on background samples,
  in: ACCV, Springer, 2009, pp. 88--97.

\bibitem{he2013half}
R.~He, W.-S. Zheng, T.~Tan, Z.~Sun, Half-quadratic-based iterative minimization
  for robust sparse representation, IEEE Trans. Pattern Anal. Mach. Intell.
  36~(2) (2013) 261--275.

\end{thebibliography}






\end{document}